\documentclass[11pt,a4paper]{article}
\usepackage[table,dvipsnames]{xcolor} % Needs to be here to avoid clashes with other packages
\usepackage[hyperref]{emnlp2020}

\usepackage{times}
\usepackage{latexsym}
\usepackage{adjustbox}
\usepackage{comment}
\usepackage{url}

\aclfinalcopy % Uncomment this line for the final submission
\usepackage{amsmath}
\usepackage{amsthm}
\usepackage{amsfonts}
\usepackage{color}
\usepackage{enumitem}
\usepackage{xspace}
\usepackage{booktabs,colortbl}  
\usepackage{array}
\usepackage{subfig}
\usepackage{bbding}
\usepackage{graphics}

\newcommand\XLMR{\mbox{XLM-R}\xspace}
\newcommand\XLMRb{\mbox{XLM-R\textsubscript{Base}}\xspace}

\newcommand\ourdataset{$E{\chi}\alpha{\mu}s$\xspace}

\newcommand\dataurl{\url{http://github.com/mhardalov/exams-qa}\xspace}

\newcommand\questionscntround{24,000\xspace}
\newcommand\questionscnt{24,143\xspace}
\newcommand\langscnt{16\xspace}
\newcommand\langgroupscnt{8\xspace}
\newcommand\subjectscnt{24\xspace}

\newcommand\devname{Dev\xspace}

\newcounter{todonumber}
\newcommand{\note}[2][]{{%
 \let\marginpar\marginnote%
 \ifodd\value{todonumber}%
   \reversemarginpar%
 \else%
 \fi%
 \todo[#1]{#2}}%
 \stepcounter{todonumber}%
}

\usepackage{color,soul}
\setul{0.5ex}{0.4ex}

\newcommand{\tc}[2]{\setulcolor{#1}\ul{#2}\setulcolor{black}}

\newcommand{\tcB}[1]{\tc{orange}{#1}}

\definecolor{lightsilver}{rgb}{0.85, 0.85, 0.85}
\newcommand{\pq}[1]{\tcB{#1}} % paralell question available
\definecolor{FancyGreen}{rgb}{0.13, 0.75, 0.42} % #20bf6b
\definecolor{FancyRed}{rgb}{0.92, 0.23, 0.35} % #eb3b5a
\definecolor{FancyOrange}{rgb}{0.98, 0.51, 0.19} % #fa8231
\newcommand{\hi}[1][0]{\cellcolor{FancyGreen!#1}}
\newcommand{\hineg}[1][0]{\cellcolor{FancyRed!#1}}
\newcommand{\sgrnew}[1]{#1\textsuperscript{*}} % same language 

\title{EXAMS: A Multi-Subject High School Examinations Dataset\\ for Cross-Lingual and Multilingual Question Answering}

\author{Momchil Hardalov$^1$ \quad Todor Mihaylov$^2$ \quad Dimitrina Zlatkova$^1$ \quad Yoan Dinkov$^1$ \\
\textbf{Ivan Koychev$^1$ \quad Preslav Nakov$^3$} \\
  $^1$Sofia University ``St. Kliment Ohridski'', Bulgaria, \\
  $^2$Heidelberg University, Germany \\
  $^3$Qatar Computing Research Institute, HBKU, Qatar, \\
  
  {\tt \{hardalov, koychev\}@fmi.uni-sofia.bg}\\
  {\tt \{dvzlatkova, jdinkov\}@uni-sofia.bg} \\
  {\tt tbmihaylov@gmail.com},
  {\tt pnakov@hbku.edu.qa}
}

\date{}

\begin{document}

\maketitle

\begin{abstract}
We propose \ourdataset~-- a new benchmark dataset for cross-lingual and multilingual question answering for high school examinations. 
We collected more than \questionscntround high-quality high school exam questions in \langscnt languages, covering \langgroupscnt language families and \subjectscnt school subjects from Natural Sciences and Social Sciences, among others. 

\ourdataset offers a fine-grained evaluation framework across multiple languages and subjects, which allows precise analysis and comparison of various models. 
We perform various experiments with existing top-performing multilingual pre-trained models and we show that \ourdataset offers multiple challenges that require multilingual knowledge and reasoning in multiple domains.
We hope that \ourdataset will enable researchers to explore challenging reasoning and knowledge transfer methods and pre-trained models for school question answering in various languages which was not possible before. 
The data, code, pre-trained models, and evaluation are available at \dataurl.
\end{abstract}

\section{Introduction}
\label{sec:intro}

Research on science question answering has attracted a lot of attention in recent years~\citep{clark2015elementary-science,Schoenick2017ai2-turing,Clark2019AristoStory}. Such questions are challenging as they require domain and common sense knowledge~\citep{Clark2018ThinkYH}, as well as complex reasoning and different forms of inference over a variety of knowledge sources~\citep{Khashabi2016QuestionAV,khashabi2018question}. Indeed, a combination of these was required to achieve noticeable performance gains~\cite{Clark2016Retrieval}. 
This inevitably made research in school-level science Question Answering (QA) hard for languages other than English due to the scarceness of resources~\citep{Clark2014automatic-kb,Khot2017sci-ie,Khot2018SciTaiLAT,Bhakthavatsalam2020GenericsKBAK}. 

There has been a recent mini-revolution in QA, as well as in the field of Natural Language Processing (NLP) in general, due to the invention of the  Transformer~\citep{NIPS2017_7181:transformer}, and the subsequent rise of large-scale pre-trained models~\citep{Peters:2018:ELMo,radford2018gpt1,radford2019language,devlin2019bert,lan2020albert,yang2019xlnet,liu2019roberta,Raffel2020ExploringTTT_T5}. 
Nowadays, fine-tuning such models on task-specific data has become an essential element of any top-scoring QA system.
Yet, for science QA, training on datasets from a different domain~\citep{sun2019readingstrategies,khashabi2020unifiedqa} and carefully selected background knowledge~\citep{Banerjee2019CarefulSO,ni-etal-2019-learning} could improve such models further. 
\begin{figure}[t!]
    \centering
    \includegraphics[width=0.45\textwidth]{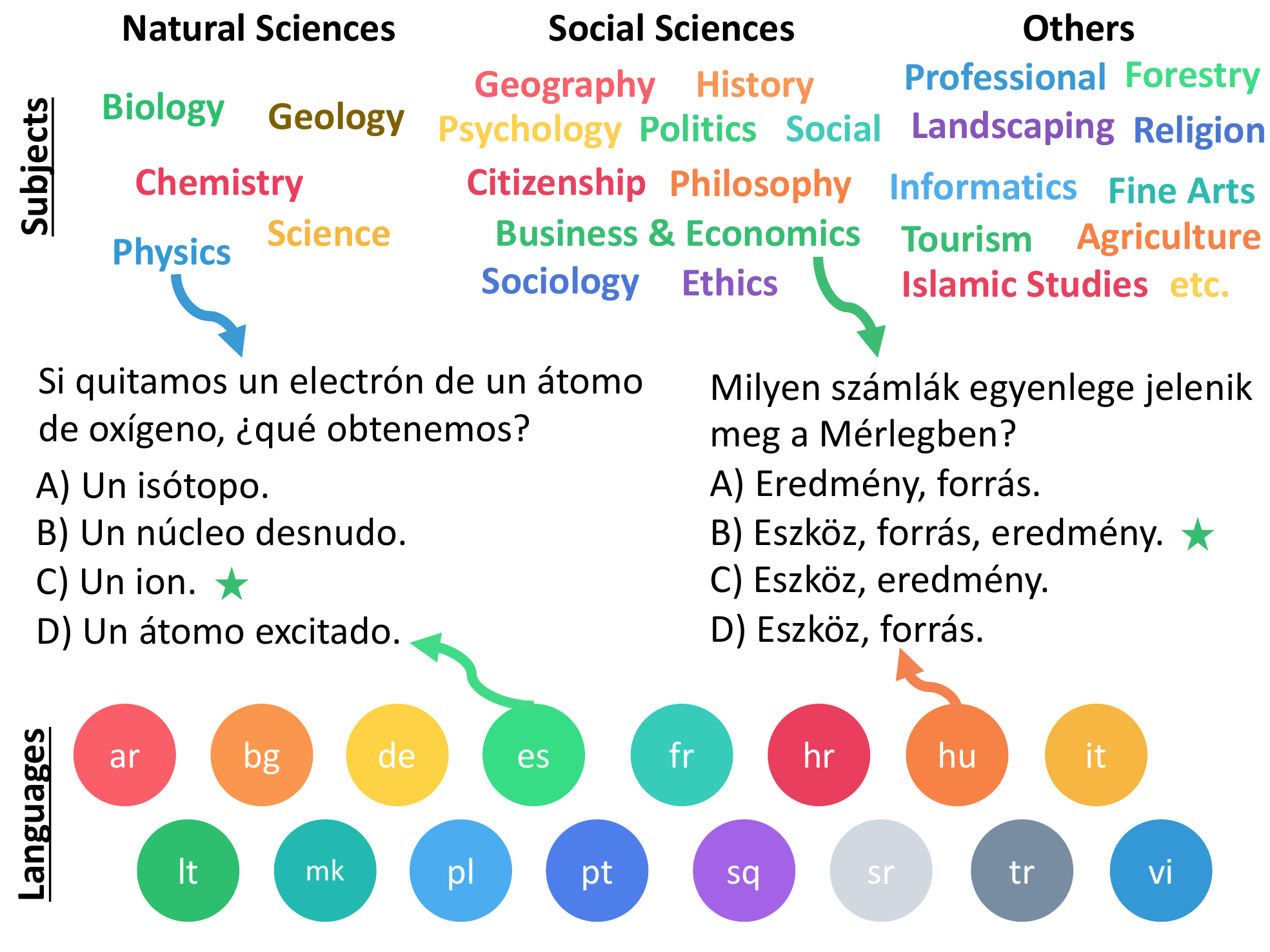}
    \caption{Properties and examples from \ourdataset. }
    \label{fig:example}
\end{figure}{}

The success of large-scale pre-trained models and the development of their multilingual versions \citep{devlin2019bert,conneau2020-xlm-roberta} gives hopes for supposedly better performance in multilingual question answering. Therefore, several new datasets have been released for multilingual reading comprehension and open-domain question answering in the Wikipedia domain \citep{liu-etal-2019-xqa,lewis-etal-2020-mlqa,Artetxe:etal:2020-xquad,clark2020tydi}.

Here, we present \ourdataset, a new dataset and benchmark for multilingual and cross-lingual evaluation of models and methods for answering diverse school science questions (see Figure~\ref{fig:example}).

Our contributions are as follows:
\begin{itemize}[nosep]
    \item We advance the task of science Question Answering (QA) with multilingual and cross-lingual evaluations.
    \item We collect a new challenging dataset \ourdataset
    \xspace from multilingual high school examinations, which offers several advantages over existing datasets:
    (\emph{i})~it covers various domains, 
    (\emph{ii})~it is nearly three times larger than pre-existing Science QA datasets,
    (\emph{iii})~it extends multilingual QA tasks to more languages, 
    (\emph{iv})~the questions are written by experts, rather than translated or crowdsourced,
    (\emph{v})~the questions are harder since they are from matriculation exams rather than 4-8th grade.
    \item We use fine-grained evaluation -- per subject and per language -- which yields more precise comparison between models.
    \item We perform extensive experiments and analysis using top-performing multilingual models (mBERT, \XLMR), and we show that \ourdataset offers several challenges that such models would need to overcome in the future, including multi-lingual and cross-lingual knowledge retrieval, aggregation, and reasoning, among others.
\end{itemize}
We release our code, pre-trained models and data for research purposes.\footnote{The \ourdataset dataset and code are publicly available at \dataurl}
\section{Related Work}
\label{sec:related_work}

\paragraph{Science QA}  
The work in Science Question Answering emerged in recent years with the development of several challenging datasets. The most notable is ARC~\citep{Clark2018ThinkYH}, which is a QA reasoning challenge that contains both \textit{Easy} and \textit{Challenge} questions from 4th to 8th grade examinations in the \textit{Natural Science} domain. As in \ourdataset, the questions in ARC are created by experts, albeit our dataset covers a wide variety of high school (8th-12th grade) subjects including but not limited to, Natural Sciences, Social Sciences, Applied Studies, Arts, Religion, etc. (see  Section~\ref{subsec:subjects} for details).
We provide definitions of the less known subjects in \ourdataset in Appendix~\ref{appendix:subject:definitions}.
\begin{table*}[t!]
\setlength{\tabcolsep}{5pt} % Default value: 6pt

\centering
    \resizebox{0.66\textwidth}{!}{%
    \begin{tabular}{llrrrrrr}
    \toprule
   \bf Lang & \bf Family &  \bf \#Subjects & \bf Question Len & \bf Choice Len & \bf \#Choices & \bf \#Questions & \bf Vocab \\
   \midrule
    Albanian & Albanian & 8 & 15.0 & 5.0 & 4.0 & 1,505 & 11,572 \\
    Arabic & Semitic & 5 & 10.3 & 3.4 & 4.0 & 562 & 5,189 \\
    Bulgarian & Balto-Slavic & 6 & 13.0 & 3.3 & 4.0 & 2,937 & 15,127 \\
    Croatian & Balto-Slavic & 14 & 14.7 & 4.1 & 3.9 & 2,879 & 20,689 \\
    French & Romance & 3 & 18.4 & 10.5 & 3.5 & 318 & 2,576 \\
    German & Germanic & 5 & 18.3 & 9.1 & 3.5 & 577 & 4,664 \\
    Hungarian & Finno-Ugric & 10 & 11.6 & 5.9 & 3.9 & 2,267 & 15,045 \\
    Italian & Romance & 12 & 20.0 & 5.6 & 3.9 & 1,256 & 9,050 \\
    Lithuanian & Balto-Slavic & 2 & 9.7 & 4.7 & 4.0 & 593 & 5,394 \\
    Macedonian & Balto-Slavic & 8 & 13.4 & 4.5 & 4.0 & 2,075 & 13,114 \\
    Polish & Balto-Slavic & 1 & 13.7 & 4.3 & 4.0 & 1,971 & 18,990 \\
    Portuguese & Romance & 4 & 19.9 & 8.6 & 4.0 & 924 & 6,811 \\
    Serbian & Balto-Slavic & 14 & 15.4 & 4.3 & 3.9 & 1,637 & 15,509 \\
    Spanish & Romance & 2 & 23.0 & 10.2 & 3.2 & 235 & 2,130 \\
    Turkish & Turkic & 8 & 19.5 & 4.6 & 4.4 & 1,964 & 22,069 \\
    Vietnamese & Austroasian & 6 & 37.0 & 6.4 & 4.0 & 2,443 & 6,076 \\
    \midrule
    \#Langs 16 & \#Families 8 & 24 & 17.19 & 5.08 & 3.96 & 24,143 & 158,942 \\
    \bottomrule
    \end{tabular}
    }
\caption{Statistics about \ourdataset. The average length of the question (\textit{{Question} Len}) and the choices (\textit{{Choice} Len}) are measured in number of tokens, and the vocabulary size (\textit{Vocab}) is measured in number of words. }
\label{tab:langquestionstats}
\end{table*}

The early versions of ARC \citep{clark2015elementary-science,Schoenick2017ai2-turing} inspired several crowdsourced datasets:
\citet{Welbl2017SciQ} proposed a scalable approach for crowdsourcing science questions given a set of basic supporting science facts.
\citet{Mishra2019ProPara} focused on specific phenomena including understanding science procedural texts,
\citet{mihaylov-etal-2018-suit} and \citet{Khot2020QASCAD} studied multi-step reasoning, given a set of science facts and commonsense knowledge,
\citet{Tafjord2019QuaRelAD}, and ~\citet{Mitra2019DeclarativeQA} worked on reasoning about qualitative relationships, and declarative texts, among others.
Unlike these English-only datasets, \ourdataset offers questions in 16 languages. Moreover, it contains questions about multiple subjects, which are presumably harder as they were extracted mostly from matriculation examinations (8-12th grade). Finally, \ourdataset contains over 24,000 questions, which is more than three times as many as in ARC.

\paragraph{Multilingual and Cross-lingual QA}
Recently, several QA datasets have been created that cover languages other than English, but still focusing on one such language.
\citet{Gupta2018MMQAAM} proposed a parallel QA task for English and Hindi,
\citet{liu2019xcmrc} collected a bilingual cloze-style dataset in Chinese and English.
\citet{jing-etal-2019-bipar} crowdsourced parallel paragraphs from novels in Chinese and English.
A few datasets investigated multiple-choice school QA~\citep{hardalov-etal-2019-beyond,nguyen2020viet-qa-dataset}, albeit in a limited domain, and for lower school grades (1st-5th). 
Other efforts focused on building bi-lingual datasets that are similar in spirit to SQuAD~\citep{Rajpurkar2016SQuAD10} -- extractive reading comprehension over open-domain articles. 
Such datasets are collected by crowdsourcing questions, following a procedure similar to~\citep{Rajpurkar2016SQuAD10}, in Russian~\citep{efimov2020sberquad}, Korean~\citep{lim2019-korquad1}, French~\citep{dhoffschmidt2020-fquad}, or by translating existing English QA pairs to Spanish~\citep{carrino2020-spanishquad}.

Recently, some multilingual datasets, were released to the public.
MLQA~\citep{lewis-etal-2020-mlqa}, and XQuAD~\citep{Artetxe:etal:2020-xquad} use translations by professionals and extend the monolingual SQuAD~\citep{Rajpurkar2016SQuAD10} to 7 and 11 languages, respectively, thus forming cross-lingual evaluation benchmarks. \citet{clark2020tydi} collected an entirely new dataset (TyDi QA) of questions in 11 typologically diverse languages. 

The task was to ask a question, and then the shortest span answering it from a list of paragraphs was selected. As these datasets are complementary, rather than making each other obsolete, hereby the recently released XTREME~\citep{hu2020xtreme} benchmark combined them in a joint task.
\ourdataset differs from the aforementioned multilingual benchmarks in several aspects.
First, we extend the multilingual QA efforts to a different, more challenging domain \citep{Clark2018ThinkYH}. Second, our datasets support more languages. Next, the questions in \ourdataset are written by educational experts rather than non-expert annotators, making the evaluation results comparable to a top-performing student.
Finally, our fine-grained evaluation for different subjects, languages, and combinations thereof allows for in-depth analysis and comparison.

\section{\ourdataset Dataset}
\label{sec:dataset}

We introduce \ourdataset, a new benchmark dataset for multilingual and cross-lingual question answering from high school examinations. 
In this section, we present the properties of the  dataset, and we give details about the process of data collection, preparation and normalization, as well as information about the data splits, and the parallel questions.

\subsection{Dataset Statistics} 
\label{subsec:datastats}
We collected \ourdataset from official state exams prepared by the ministries of education of various countries. 
These exams are taken by students graduating from high school,
and often require knowledge learned through the entire course.
The~questions cover a large variety of subjects and material based on the country's education system. 
Moreover, we do not focus only on major school subjects such as Biology, Chemistry, Geography, History, and Physics, but we also cover highly-specialized ones such as Agriculture, Geology, Informatics, as well as some applied and profiled studies. 
These characteristics make the questions in the dataset of very high variety, and not easily solvable, due to the need for highly specialized knowledge. 
Next, we discuss the cross-lingual and the multilingual properties of our dataset.

\paragraph{Parallel Questions}
\begin{table}[t!]
    \centering
    \setlength{\tabcolsep}{3.7pt}
    \resizebox{0.42\textwidth}{!}{%
    \begin{tabular}{cccccccccc}
    \toprule
    {} & \textbf{de} & \textbf{es} & \textbf{fr} & \textbf{hr} & \textbf{hu} & \textbf{it} & \textbf{mk} & \textbf{sq} & \textbf{sr} \\
    \midrule
    \textbf{de} &           - &             &             &             &             &             &             &             &             \\
    \textbf{es} &         199 &           - &             &             &             &             &             &             &             \\
    \textbf{fr} &         253 &         120 &           - &             &             &             &             &             &             \\
    \textbf{hr} &         189 &         134 &         109 &           - &             &             &             &             &             \\
    \textbf{hu} &         456 &         159 &         274 &         236 &           - &             &             &             &             \\
    \textbf{it} &          30 &           9 &          15 &        1,214 &          99 &           - &             &             &             \\
    \textbf{mk} &           0 &           0 &           0 &           0 &           0 &           0 &           - &             &             \\
    \textbf{sq} &           0 &           0 &           0 &           0 &           0 &           0 &        1,403 &           - &             \\
    \textbf{sr} &          40 &          25 &          20 &        1,564 &         104 &        1,002 &           0 &           0 &           - \\
    \textbf{tr} &           0 &           0 &           0 &           0 &           0 &           0 &        1,222 &         981 &           0 \\
    \bottomrule
    \end{tabular}
}
    \caption{Parallel questions for different language pairs.}
    \label{tab:parallel}
\end{table}

Some countries allow students to take official examinations in several languages. Such parallel examinations also exist in our dataset.
In particular, there are 9,857 parallel question pairs spread across seven languages as shown in Table~\ref{tab:parallel}. 
The parallel pairs are coming from Croatia (Croatian, Serbian, Italian, Hungarian), Hungary (Hungarian, German, French, Spanish, Croatian, Serbian, Italian), and North Macedonia (Macedonian, Albanian, Turkish).

\paragraph{Multilinguality}

Our dataset includes a total of \questionscnt questions in 16 languages from eight language families. Each question is a 3-way to 5-way (3.96 on average) multiple-choice question with a single correct answer. Table~\ref{tab:langquestionstats} shows a breakdown for each language, where the number of subjects, questions, and the vocabulary size are shown as absolute numbers, while the question length, the choice length, and the number of choices are averaged. All statistics about the questions and the answer options are measured in terms of words. We see that we have a rich vocabulary with almost 160,000 unique words. Interestingly, there are $\sim$9,500 shared words between at least one pair of languages in our dataset, excluding numbers and punctuation. 
As expected, the overlapping words are mostly between closely related languages (bg-mk, bg-sr, es-it, es-pt, hr-sr, mk-sr).
Other common shared words are subject-specific words such as person names (e.g.,~\emph{Abraham}, \emph{Karl}, \emph{Ivan}), chemical compounds (e.g.,~\emph{NaOH}, \emph{HCl}), units (e.g.,~\emph{m/s}, \emph{g/mol}), etc. 
Then, there are cognates with the exact same spelling (homographs) even between unrelated languages, mostly words of Latin or Greek origin, e.g.,~\textit{temperatura} (temperature) and \textit{forma} (form).
Finally, there are also \emph{false friends}, whose meaning differs across languages, e.g.,~\textit{para} can mean \emph{for} (es/pt) vs. \emph{money} (mk/tr/sq) vs. \emph{couple} (pl); similarly, \textit{ser} can mean \emph{be} (es/pt) vs. \emph{cheese} (pl) vs. \emph{after} (vi).

\subsection{Subjects and Categories}
\label{subsec:subjects}

\begin{figure}[t!]
    \centering
    \includegraphics[width=0.45\textwidth]{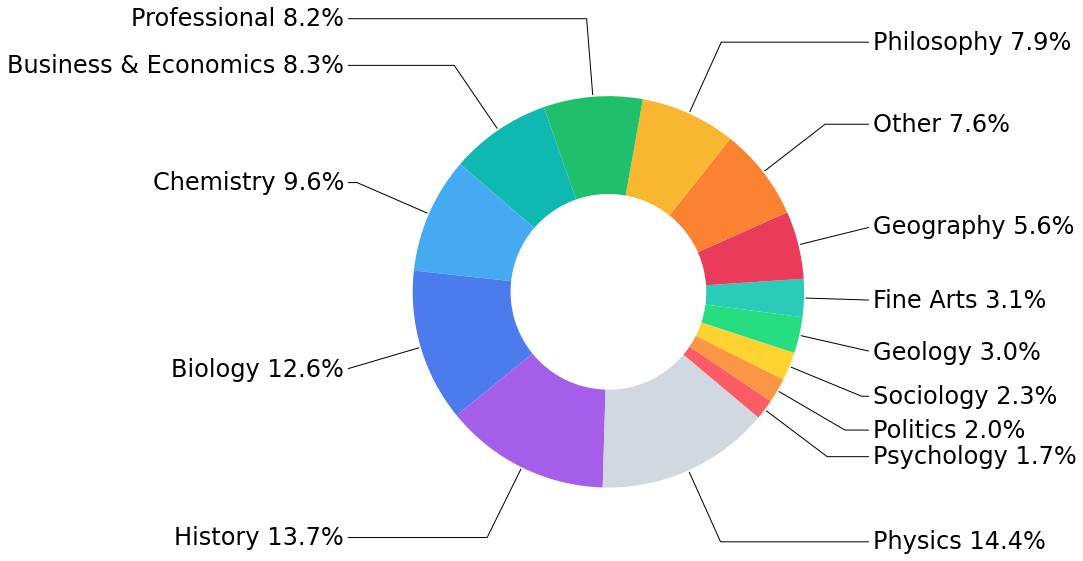}
    \caption{Relative sizes of the subjects. Those that cover less than 1.5\% of the examples are in \textit{Other}.}
    \label{fig:subject_dist}
\end{figure}{}

Each education system has its own specifics, resulting in some differences in curricula, topics, and even naming of the subjects. That being said, the original, non-normalized categories in our dataset are more than 40 for exams from just a few countries. Given the sparse nature of the subjects, we use a two-level taxonomy in order to categorize them into logically connected groups. 
The lower-level is a subject, and the higher level is a major group. We normalized the subject using a two-step algorithm: first, we put each subject (with its original naming) in a separate category, then, if the subject was general enough, e.g.,~Biology, History, etc., or there were no similar ones, we retained the category; otherwise, we merged all similar subjects together in a unifying category, e.g.,~Economics Basics, and Economics \& Marketing. We repeated the aforementioned steps until there were no suitable merge candidates. As a result, we ended up with a total of \subjectscnt subjects (see Appendix~\ref{sec:subject_analysis} for more details), which we further grouped into three major categories, based on the main branches of science: \textbf{Natural Science} -- \emph{``the study of natural phenomena''}, \textbf{Social Sciences} -- \emph{``the study of human behavior and societies''}, \textbf{Other} -- \textit{Applied Studies, Arts, Religion, etc.} (see Figure \ref{fig:example}). \footnote{\href{https://en.wikipedia.org/wiki/Branches\_of\_science}{https://en.wikipedia.org/wiki/Branches\_of\_science}}

The distribution of the major categories is \textit{Natural Sciences} 
(40.0\%) 
and \textit{Social Sciences} 
(44.0\%) and 16.0\% for \textit{Others} 
(these are the actual numbers, not approximate).
The remaining questions are labeled as \textit{Other} as they are not suitable for the two main categories.
Figure~\ref{fig:subject_dist} presents the relative sizes of the subjects in the dataset.

\subsection{Collection and Preparation}
\label{subsec:collection}
Here, we describe the process of collecting and preparing the data, as it is not trivial and it could be applied to other languages and  examinations.
First, we identified potential online sources of publicly available school exams starting from the \emph{Matriculation Examination} page in Wikipedia.\footnote{
\href{https://en.wikipedia.org/wiki/Matriculation\_examination}{https://en.wikipedia.org/wiki/Matriculation\_examination}}

For all languages in our dataset, the first step in the process of data collection was to download the PDF files per year, per subject, and per language (when parallel languages were available in the same source). 
We converted the PDF files to text and we used only those that were well-formatted and followed the document structure.
 
Then, we used Regular Expressions (RegEx) to parse the questions, their corresponding choices and the correct answer choice.
In order to ensure that all our questions are answerable using textual input only, we removed questions that contained visual information. 
We did that using a manually curated list of words such as \textit{map}, \textit{table}, \textit{picture}, \textit{graph}, etc., in the corresponding language. Next,~we performed data cleaning to ensure the quality of the generated dataset, by manually reviewing each question and its choices and ensuring that all options, text, and symbols (e.g.,~$\mu, \rightarrow, \alpha, \leftarrow$) were displayed correctly. 
As a result, we filtered out about 17\% of the questions (the percentage varies based on the source, the language, and the subject).
Finally, in order to remove frequency bias such as ``most answers are B)'', we shuffled each question's choices.

\subsection{Data Splits}
\label{subsec:datasplits}

\begin{table}[t!]
\small
    \centering
    \resizebox{0.5\textwidth}{!}{%
    \begin{tabular}{@{}l|r@{ }r@{ }r|rr@{ }}
    \toprule
    {} & \multicolumn{3}{c|}{\bf Multilingual} & \multicolumn{2}{c}{\bf Cross-lingual} \\
    % \midrule
    \bf Language & \multicolumn{1}{c}{\bf Train} &  \multicolumn{1}{c}{\bf Dev} & \multicolumn{1}{c|}{\bf Test} & \multicolumn{1}{c}{\bf Train} & \multicolumn{1}{c}{\bf Dev} \\
    \midrule
    Albanian         &         565 &    185 &   755 &        1,194 &  311 \\
    Arabic           &           - &      - &   562 &            - &  -   \\
    Bulgarian        &       1,100 &    365 & 1,472 &        2,344 &  593 \\
    Croatian         &       1,003 &    335 & 1,541 &        2,341 & 538  \\
    French           &           - &      - &   318 &            - &  -   \\
    German           &           - &      - &   577 &            - &    - \\
    Hungarian        &         707 &    263 & 1,297 &        1,731 & 536  \\
    Italian          &         464 &    156 &   636 &        1,010 & 246  \\
    Lithuanian       &           - &      - &   593 &            - &    - \\
    Macedonian    &         778 &    265 & 1,032 &           1,665 & 410  \\
    Polish           &         739 &    246 &   986 &        1,577 & 394  \\
    Portuguese       &         346 &    115 &   463 &          740 & 184    \\
    Serbian          &         596 &    197 &   844 &        1,323 & 314  \\
    Spanish          &           - &      - &   235 &            - &    - \\
    Turkish          &         747 &    240 &   977 &        1,571 & 393  \\
    Vietnamese       &         916 &    305 & 1,222 &        1,955 & 488   \\\midrule
    Combined       &   7,961 &    2,672 & 13,510 &        - & -   \\
    \bottomrule
    \end{tabular}
    }
    \caption{Number of examples in the data splits based on the experimental setup. 
    }
    \label{tab:datasplits}
\end{table}

In our experiments, we aim at evaluating the multilingual and the cross-lingual question answering capabilities of different models. Therefore, we split the data in order to support both evaluation strategies: \textit{Multilingual} and \textit{Cross-lingual}.

\paragraph{Multilingual} In this setup, we want to train and to evaluate a given model with multiple languages, and thus we need multilingual \textit{training}, \textit{validation} and \textit{test} sets. 
In order to ensure that we include as many of the languages as possible, we first split the questions independently for each language $L$ into Train$_L$, Dev$_L$, Test$_L$ with 37.5\%, 12.5\%, 50\% of the examples, respectively.\footnote{For languages with fewer than 900 examples, we only have Test$_L$.}
We then unite all language-specific subsets into the multilingual sets Train$_{Mul}$, Dev$_{Mul}$, Test$_{Mul}$, and we used them for training, development, and testing.

Since we have parallel data for several languages (discussed in Section~\ref{subsec:datastats}), in this setup, we ensure that the same parallel questions are only found in either training, development or testing, so that we do not leak the answer from training via some other language. In order to do that, we sample the questions with the assumptions and the ratios mentioned above, stratified per subject in the given language. The number of examples per language and the total number of multilingual sets are shown in the first three columns of Table~\ref{tab:datasplits}.\footnote{Sometimes, grouping parallel questions in the same split slightly violates the splitting ratios.}

\paragraph{Cross-Lingual} In this setting, we want to explore the capability of a model to transfer its knowledge from a single source language~$L_{src}$ to a new unseen target language~$L_{tgt}$. 
In order to ensure that we have a larger training set, we train the model on 80\% of $L_{src}$, we validate on 20\% of the same language, and we test on a subset of $L_{tgt}$.\footnote{To ensure that the cross-lingual evaluation is comparable to the multilingual one, we use the same subset of questions from language $L_{tgt}$ that are used in Test$_{Mul}$}
The last three columns of Table \ref{tab:datasplits} show the number of examples used for training and validation with the corresponding language. 

\subsection{Reasoning and Knowledge Types}
\label{sec:reasoning_types}

\begin{figure}[t!]
    \centering
    \includegraphics[width=\columnwidth]{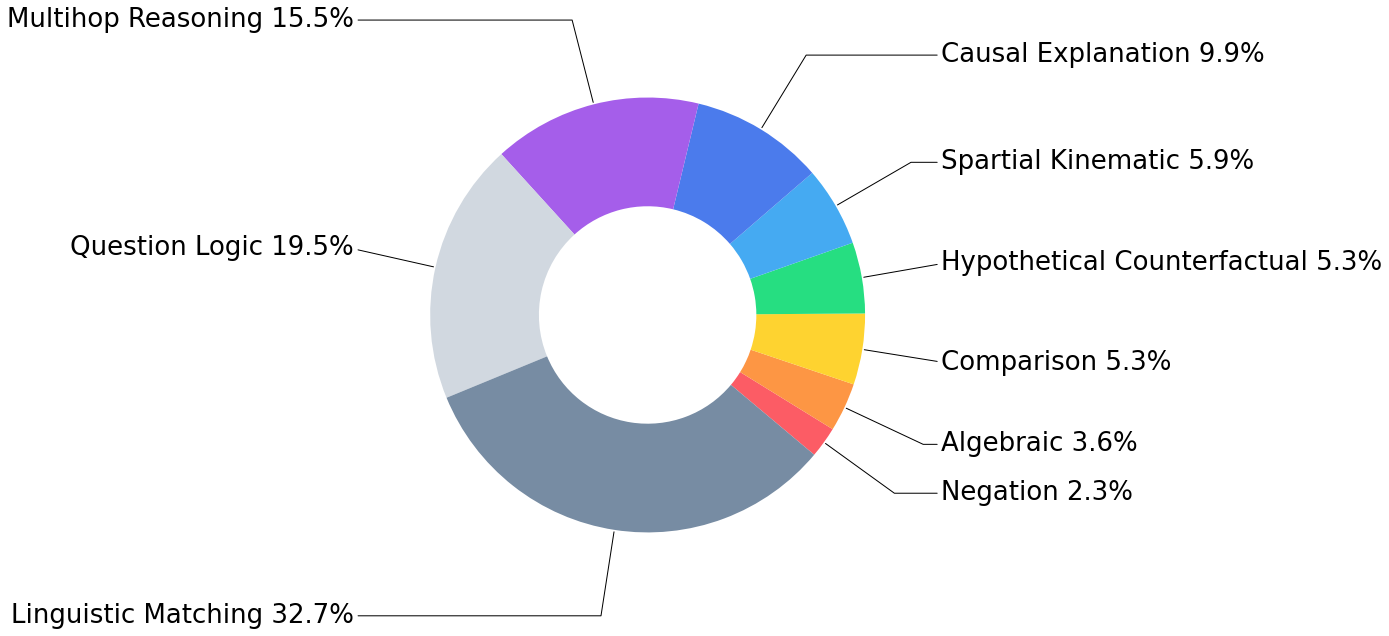}
    \caption{Relative sizes of reasoning types in \ourdataset.}
    \label{fig:reason_dist}
\end{figure}

In order to give a better understanding of the reasoning, and the knowledge types in \ourdataset, we sampled and annotated 250 questions, all of which are from the multilingual \devname.
For each question, we provided English translations
as not all annotators were native speakers of the questions' language. We followed the procedure and re-used the annotation types presented in earlier work~\citep{Clark2018ThinkYH,Boratko2018-sciq-annotations}. However, as they were designed mainly for Nature Science questions, we extended them with two new annotation types: \textit{``Domain Facts and Knowledge''} and \textit{``Negation''} (see Appendix~\ref{appendix:reasoning_types} for examples).

\begin{figure}[t!]
    \centering
    \includegraphics[width=\columnwidth]{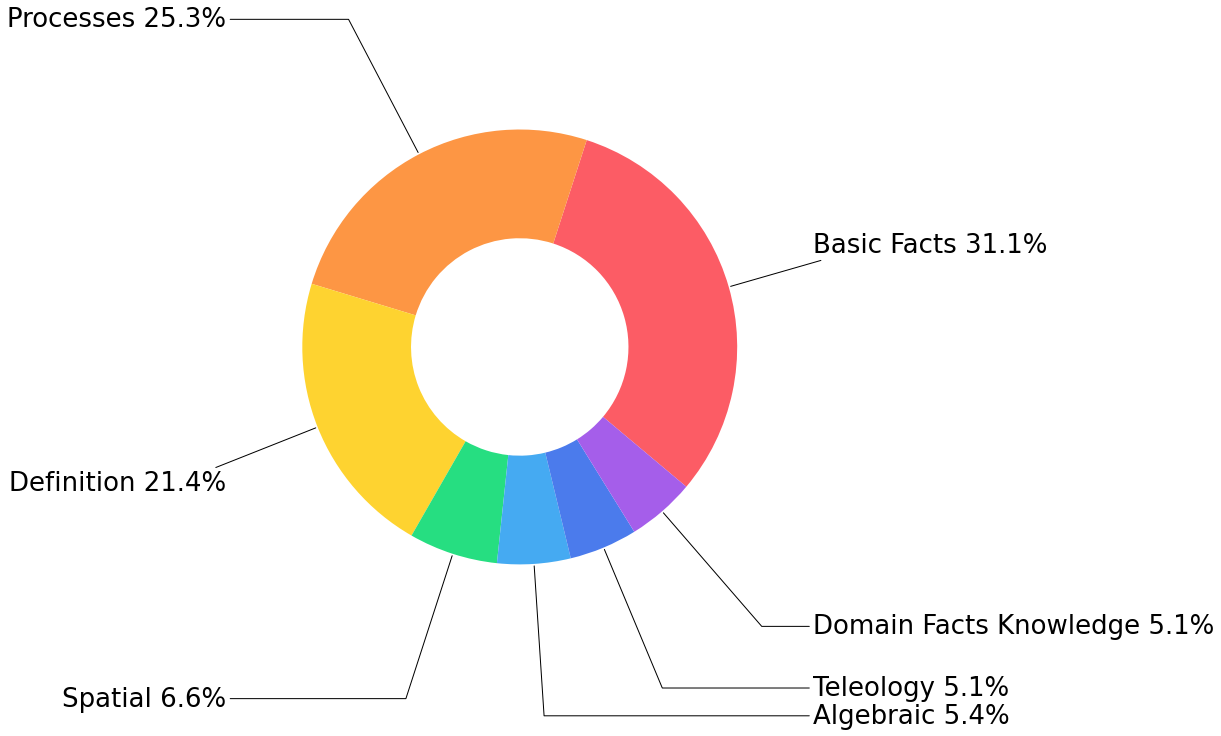}
    \caption{Relative size of the \ourdataset knowledge types.}
    \label{fig:knowledge_dist}
\end{figure}

The relative sizes of the knowledge and the reasoning types are shown in Figures~\ref{fig:reason_dist} and \ref{fig:knowledge_dist}. Here, we must note that the sizes are approximate rather than exact, since the annotations are subjective and the distribution may vary.

\section{Baseline Models}
\label{sec:models}

We divide our baselines into the following two categories: (\textit{i})~models without additional training, and (\textit{ii})~fine-tuned models. The first group contains common baselines, i.e.,~random guessing and information retrieval solver~\citep{Clark2016Retrieval}. In addition, we evaluate the knowledge contained in the pre-trained language model, i.e.,~mBERT~\citep{devlin2019bert} and XLM-R~\citep{conneau2020-xlm-roberta}, and we use it as an answering mechanism. The second group of baselines compare the learning ability of state-of-the-art multilingual models on the task of multiple-choice question answering. Since we have multi-choice questions, we adopt accuracy as an evaluation measure, as this is standard for this setup.

\subsection{No Additional Training}
\label{subsec:no_training}

\paragraph{Information Retrieval (IR)}

This IR baseline is from \citet{Clark2016Retrieval}, and it ranks the possible options $o$ for each question $q$ based on the relevance score returned by a search engine\footnote{We build and use a separate index for each language using ElasticSearch.}. In particular, for each option $o_i$, we form a query by appending the option's text to the question's ($q + o_i$), and we send this concatenation to the search engine. We then sum the returned scores for the top-10 hits, and we predict the choice with the highest score to be the correct answer. 
More detailed discussion can be found in Appendix~\ref{sec:wikicorpus}.

\begin{table*}[t!]
\setlength{\tabcolsep}{3pt} % Default 
\resizebox{\textwidth}{!}{%
\renewcommand{\arraystretch}{1.1}
\begin{tabular}{l|ccc|cccccccccccccccc|c}
\toprule
 & \multicolumn{2}{c}{\bf ARC} & \bf R12 & \multicolumn{17}{c}{\bf \ourdataset} \\
\bf Lang/Set & \bf E & \bf C & \bf en & \bf ar & \bf bg & \bf de & \bf es & \bf fr & \bf hr &
\bf hu & \bf it & \bf lt & \bf mk & \bf pl & \bf pt & \bf sq & \bf sr & \bf tr & \bf vi & \bf All \\ 
\midrule
Random Guess                   &      25.0 &      25.0 &  25.0 &  25.0 &  25.0 &  29.4 &  32.0 &  29.4 &  26.7 &  27.7 &  26.0 &  25.0 &  25.0 &  25.0 &  25.0 &  25.0 &  26.2 &  23.1 &  25.0 & 25.9 \\
IR (Wikipedia)                       &      - &      - &   - &  31.0 &  29.6 &  29.3 &  27.2 &  32.1 &  31.9 &  29.7 &  27.6 &  29.8 &  32.2 &  29.2 &  27.5 &  25.3 &  31.8 &  28.5 &  27.5 & 29.5 \\
\midrule
\XLMR on RACE         &     61.6 &     45.9 &  57.4 &  39.1 &  43.9 &  37.2 &  40.0 &  37.4 &  38.8 &  39.9 &  36.9 &  40.5 &  45.9 &  33.9 &  37.4 &  42.3 &  35.6 &  37.1 &  35.9 &   39.1 \\
% then on 
w/ SciENs              &     \bf73.6 &     51.2 &  68.4 &  39.1 &  44.2 &  35.5 &  37.9 &  37.1 &  38.5 &  37.9 &  39.5 &  \bf41.3 &  49.8 &  36.1 &  \bf39.3 &  42.5 &  37.4 &  37.4 &  35.9 &   39.6 \\
then on \ourdataset (Full) &    72.8 &     \bf52.6 &  \bf68.8 &  \bf40.7 &  \bf47.2 &  \bf39.7 &  \bf42.1 &  \bf39.6 &  \bf41.6 &  \bf40.2 &  \bf40.6 &  40.6 &  \bf53.1 &  \bf38.3 &  38.9 &  \bf44.6 &  \bf39.6 &  \bf40.3 &  \bf37.5 &   \bf42.0\\
\midrule
\XLMRb (Full) &     54.2 &     36.4 &  54.6 &  34.5 &  35.7 &  \it36.7 &  38.3 &  \it36.5 &  \it35.6 &  \it33.3 &  33.3 &  33.2 &  41.4 &  \it30.8 &  29.8 &  \it33.5 &  32.3 &  30.4 &  \it32.1 &   34.1 \\
mBERT  (Full)    &        \it63.8 &     \it38.9 &  \it57.0 &  34.5 &  \it39.5 &  35.3 &  \it40.9 &  34.9 &  35.3 &  32.7 &  \it36.0 &  \it34.4 &  \it42.1 &  30.0 &  29.8 &  30.9 &  \it34.3 &  \it31.8 &  31.7 &   \it34.6 \\
mBERT (\ourdataset only)       &     39.6 &     28.5 &  35.1 &  31.9 &  34.1 &  30.4 &  37.9 &  33.3 &  32.6 &  29.3 &  31.1 &  31.9 &  42.4 &  29.0 &  28.3 &  29.9 &  30.8 &  25.4 &  30.0 & 31.7 \\
\midrule
\XLMR as KB             &     30.8 &     26.2 &  27.2 &  31.0 &  27.2 &  31.7 &  37.9 &  29.9 &  27.6 &  29.3 &  28.0 &  28.3 &  23.5 &  24.6 &  27.0 &  25.6 &  25.4 &  24.4 &  24.9 &   27.0 \\
\XLMR (Full) w/o ctx        &     45.4 &     39.2 &  47.6 &  30.2 &  34.8 &  34.3 &  30.2 &  33.0 &  33.6 &  33.4 &  28.5 &  30.9 &  37.5 &  30.0 &  32.4 &  36.7 &  32.1 &  31.7 &  30.4 &   32.8 \\
\bottomrule
\end{tabular}%
}
\caption{Overall per-language evaluation. The first three columns show the results on ARC Easy (E), ARC Challenge (C), and Regents 12 LivEnv (en). The following columns show the per-language and the overall results (the last column All) for all languages. \textit{All} is the score averaged over all \ourdataset questions.}
\label{tab:res:overall-lang}
\end{table*}

\paragraph{Pre-trained Model as a Knowledge Base (KB)} 
As we start to understand pre-trained BERT-like models better\citep{petroni2019language,rogers2020primer}, we observe some interesting phenomena. Here, we evaluate the knowledge contained in the model by leveraging the standard masking mechanism used in pre-training. We tokenize each question-option pair into subwords,
and then we replace all the pieces from the option with the special [MASK] token.
Following the notation from~\citet{devlin2019bert}, the input sequence can be written as follows:

\vspace{1mm}
\noindent{[CLS] [Q\textsubscript{1}] \dots [Q\textsubscript{N}] [M\_O\textsubscript{1}] \dots [M\_O\textsubscript{M}] [SEP]},
\vspace{1mm}

\noindent where Q is the question, and M\_O is the masked option. Following the notation above, we obtain a score for each option in the question based on the normalized log-probability for the entire masked sequence. (see Eq.~\ref{eq:bertkbscore}).
\begin{equation}
    score(O_i) = \frac{1}{|O_i|}\sum_{t \in O_i}{\log{P_{MLM}(t|Q)}}
    \label{eq:bertkbscore}
\end{equation}

We could probably obtain better results for that evaluation if we form the question-option pairs as a single statement, e.g.,~``What is the purpose of \textit{something}? [SEP] [M\_O] $\rightarrow$ The purpose of \textit{something} is [M\_O].''

\subsection{Fine-Tuned Models}
\label{subsec:fine-tuning}

We are interested in evaluating the ability of pre-trained models to transfer science-based knowledge across languages when fine-tuned. 

In order to evaluate the QA capability of these models, we follow the established approach in this setting~\citep{devlin2019bert, liu2019roberta, sun2019readingstrategies}, and we fine-tune them to predict the correct answer in a multi-choice setting, given a selected context.
This setup feeds the pre-trained model with a text, tokenized using the corresponding tokenizer for the model in the format: 

[CLS] C [SEP] Q + O [SEP],

\noindent where C, Q and O are the tokenized \textit{knowledge context} (see Appendix~\ref{sec:wikicorpus}),
the \textit{question}, and the \textit{option}, respectively.  Each question-option pair (Q+O) is evaluated, and the one with the highest confidence of being an answer is selected.

In our experiments, we used the Transformers library~\citep{Wolf2019HuggingFaces-Transformers}. 
We experimented with the best-performing multilingual models: the Multilingual version of BERT, or mBERT~\citet{devlin2019bert}, and the recently proposed XLM-RoBERTa, or \XLMR~\cite{conneau2020-xlm-roberta}.

\textbf{Multilingual BERT}~\citep{devlin2019bert} is a fundamental multilingual model trained on 104 languages with a vocabulary of 110K word-pieces, with a total of 172M parameters (12 layers, 768 hidden states, 12 heads).

\textbf{XLM-RoBERTa}~\citep{conneau2020-xlm-roberta} is a recent multilingual model based on RoBERTa~\citep{liu2019roberta}. It is trained on 100 languages, with a larger vocabulary of 250K sentence pieces. It comes in two sizes: \textit{XLM-R$_{Base}$} (270M parameters, same architecture as mBERT, except vocab size), and \textit{XLM-R} (550M parameters, 24 layers, 1,024 hidden states, 16 heads). For completeness, we include both in our experiments.

We fine-tuned the aforementioned models following the standard procedure for multiple-choice comprehension tasks, as described in~\citep{devlin2019bert} and \citep{liu2019roberta}, using the Transformers library~\citep{Wolf2019HuggingFaces-Transformers}. The training details can be found in Appendix~\ref{sec:training-and-hyper-params}.

\section{Experiments and Results}
\label{sec:experiments_and_results}

In this section, we evaluate the performance of the baseline models described in Section \ref{sec:models} on the \ourdataset dataset.
In Table \ref{tab:res:overall-lang}, we show the overall per-language performance of the evaluated models. 
The first group shows simple baselines: random guessing and IR over Wikipedia articles. IR is better than random guessing, but it is clear that most questions require reasoning beyond simple word matching.
In the last group, we evaluate the knowledge contained in the models before and after the QA fine-tuning. 
First, we evaluate \XLMR as a knowledge base, and then we use the \textit{Full} model but with the question--option pair only.

\begin{table*}[t!]
\centering
\setlength{\tabcolsep}{3pt}
\resizebox{0.95\textwidth}{!}{
\renewcommand{\arraystretch}{1.1}
\begin{tabular}{l|ccc|cccccccccccccccc}
\toprule
\bf Lang &     \bf A\textsubscript{E} &    \bf A\textsubscript{Ch} &        \bf R12 &              \bf de &                 \bf es &            \bf fr &         \bf it &        \bf pt &           \bf  bg &                \bf  hr &           \bf  lt &         \bf     mk &        \bf     pl &       \bf      sr &         \bf     hu &      \bf    sq &        \bf      tr &   \bf      vi &    \bf     ar \\
\midrule
en$_{all}$ &     \sgrnew{73.6} &     \sgrnew{51.2} &     \sgrnew{68.4} &      \sgrnew{35.5} &                            37.9 &                      37.1 &                  39.5 &                     39.3 &                       44.2 &                       38.5 &                       41.3 &                       49.8 &                  36.1 &                       37.4 &              37.9 &              42.5 &              37.4 &              35.9 &              39.1 \\
\midrule
w/ it     &       \hi[20]+1.4 &       \hi[18]+1.3 &       \hi[19]+1.4 &  \hi[60]\pq{+6.2} &  \hi[60]\sgrnew{\pq{\bf{+4.2}}} &  \hi[4]\sgrnew{\pq{+0.3}} &                     - &  \hineg[52]\sgrnew{-3.7} &                \hi[17]+1.2 &           \hi[58]\pq{+4.1} &                \hi[12]+0.9 &                \hi[11]+0.8 &           \hi[21]+1.5 &           \hi[44]\pq{+3.1} &  \hi[40]\pq{+2.8} &       \hi[12]+0.9 &    \hineg[18]-1.3 &  \hi[25]\bf{+1.8} &       \hi[25]+1.8 \\
w/ pt     &        \hi[1]+0.1 &       \hi[17]+1.2 &    \hineg[11]-0.8 &       \hi[31]+2.2 &            \hi[35]\sgrnew{+2.5} &   \hineg[35]\sgrnew{-2.5} &  \hi[19]\sgrnew{+1.4} &                        - &                 \hi[4]+0.3 &                  \hi[0]0.0 &                \hi[28]+2.0 &                \hi[11]+0.8 &         \hineg[1]-0.1 &              \hineg[8]-0.6 &     \hineg[8]-0.6 &    \hineg[18]-1.3 &  \hi[18]\bf{+1.3} &        \hi[8]+0.6 &       \hi[15]+1.1 \\
w/ bg     &        \hi[8]+0.6 &        \hi[5]+0.4 &     \hineg[5]-0.4 &  \hi[51]\bf{+3.6} &                     \hi[11]+0.8 &               \hi[22]+1.6 &      \hi[48]\bf{+3.4} &           \hineg[27]-1.9 &                          - &       \hi[21]\sgrnew{+1.5} &  \hi[41]\sgrnew{\bf{+2.9}} &  \hi[22]\sgrnew{\bf{+1.6}} &   \hi[1]\sgrnew{+0.1} &  \hi[21]\sgrnew{\bf{+1.5}} &       \hi[28]+2.0 &  \hi[32]\bf{+2.3} &    \hineg[12]-0.9 &    \hineg[11]-0.8 &       \hi[11]+0.8 \\
w/ hr     &       \hi[15]+1.1 &  \hi[24]\bf{+1.7} &     \hineg[2]-0.2 &  \hi[60]\pq{+4.8} &                \hi[54]\pq{+3.8} &           \hi[4]\pq{+0.3} &      \hi[60]\pq{+5.8} &           \hineg[39]-2.8 &       \hi[24]\sgrnew{+1.7} &                          - &        \hi[2]\sgrnew{+0.2} &     \hineg[1]\sgrnew{-0.1} &  \hi[17]\sgrnew{+1.2} &  \hi[60]\sgrnew{\pq{+6.7}} &  \hi[40]\pq{+2.8} &       \hi[24]+1.7 &       \hi[17]+1.2 &        \hi[7]+0.5 &     \hineg[1]-0.1 \\
w/ mk     &       \hi[21]+1.5 &     \hineg[7]-0.5 &  \hi[31]\bf{+2.2} &       \hi[14]+1.0 &                \hi[60]\bf{+4.2} &             \hineg[4]-0.3 &           \hi[28]+2.0 &           \hineg[37]-2.6 &       \hi[25]\sgrnew{+1.8} &  \hi[55]\sgrnew{\bf{+3.9}} &       \hi[21]\sgrnew{+1.5} &                          - &  \hi[27]\sgrnew{+1.9} &         \hi[0]\sgrnew{0.0} &       \hi[28]+2.0 &  \hi[60]\pq{+6.9} &  \hi[60]\pq{+4.8} &        \hi[7]+0.5 &  \hi[60]\bf{+4.5} \\
w/ pl     &    \hineg[28]-2.0 &    \hineg[21]-1.5 &    \hineg[44]-3.1 &         \hi[0]0.0 &                      \hi[5]+0.4 &            \hineg[35]-2.5 &            \hi[1]+0.1 &           \hineg[18]-1.3 &       \hi[15]\sgrnew{+1.1} &       \hi[14]\sgrnew{+1.0} &     \hineg[7]\sgrnew{-0.5} &     \hineg[2]\sgrnew{-0.2} &                     - &         \hi[0]\sgrnew{0.0} &     \hineg[5]-0.4 &        \hi[4]+0.3 &        \hi[2]+0.2 &    \hineg[19]-1.4 &       \hi[12]+0.9 \\
w/ sr     &  \hi[25]\bf{+1.8} &     \hineg[1]-0.1 &    \hineg[17]-1.2 &  \hi[37]\pq{+2.6} &                \hi[60]\pq{+5.1} &          \hi[27]\pq{+1.9} &      \hi[39]\pq{+2.8} &            \hineg[8]-0.6 &  \hi[31]\sgrnew{\bf{+2.2}} &  \hi[60]\sgrnew{\pq{+6.2}} &        \hi[2]\sgrnew{+0.2} &       \hi[18]\sgrnew{+1.3} &  \hi[18]\sgrnew{+1.3} &                          - &  \hi[19]\pq{+1.4} &     \hineg[5]-0.4 &     \hineg[9]-0.7 &    \hineg[14]-1.0 &       \hi[45]+3.2 \\
w/ hu     &    \hineg[11]-0.8 &    \hineg[11]-0.8 &    \hineg[14]-1.0 &  \hi[60]\pq{+7.8} &               \hi[60]\pq{+10.2} &          \hi[39]\pq{+2.8} &      \hi[15]\pq{+1.1} &           \hineg[27]-1.9 &                 \hi[9]+0.7 &           \hi[11]\pq{+0.8} &             \hineg[45]-3.2 &                 \hi[1]+0.1 &           \hi[12]+0.9 &           \hi[12]\pq{+0.9} &                 - &     \hineg[2]-0.2 &     \hineg[2]-0.2 &     \hineg[8]-0.6 &    \hineg[19]-1.4 \\
w/ sq     &     \hineg[1]-0.1 &        \hi[4]+0.3 &    \hineg[21]-1.5 &       \hi[50]+3.5 &                   \hineg[7]-0.5 &             \hineg[8]-0.6 &           \hi[11]+0.8 &              \hi[12]+0.9 &                \hi[12]+0.9 &                \hi[11]+0.8 &                \hi[14]+1.0 &           \hi[48]\pq{+3.4} &            \hi[8]+0.6 &                 \hi[8]+0.6 &       \hi[27]+1.9 &                 - &   \hi[5]\pq{+0.4} &        \hi[4]+0.3 &        \hi[2]+0.2 \\
w/ tr     &     \hineg[7]-0.5 &       \hi[15]+1.1 &    \hineg[21]-1.5 &       \hi[21]+1.5 &                     \hi[42]+3.0 &            \hineg[27]-1.9 &           \hi[32]+2.3 &           \hineg[42]-3.0 &                \hi[14]+1.0 &                \hi[14]+1.0 &             \hineg[38]-2.7 &           \hi[21]\pq{+1.5} &            \hi[2]+0.2 &                \hi[17]+1.2 &  \hi[34]\bf{+2.4} &  \hi[52]\pq{+3.7} &                 - &    \hineg[14]-1.0 &       \hi[25]+1.8 \\
w/ vi     &     \hineg[7]-0.5 &        \hi[5]+0.4 &    \hineg[11]-0.8 &       \hi[41]+2.9 &                     \hi[48]+3.4 &          \hi[58]\bf{+4.1} &           \hi[15]+1.1 &         \hi[15]\bf{+1.1} &                \hi[21]+1.5 &                \hi[24]+1.7 &                 \hi[5]+0.4 &                 \hi[5]+0.4 &      \hi[30]\bf{+2.1} &                  \hi[0]0.0 &       \hi[24]+1.7 &       \hi[11]+0.8 &       \hi[15]+1.1 &                 - &       \hi[48]+3.4 \\
\bottomrule
\end{tabular}

}
\caption{Cross-lingual zero-shot performance on \ourdataset. The first three columns show the performance on the test set of the AI2 science datasets (English), followed by per-language evaluation. The \pq{underlined} values mark languages that have parallel data with the source language, and the ones with an asterisk\textsuperscript{*} are from the same family.}
\label{tab:res:cross-ling-zeroshot}
\end{table*}

\subsection{Multilingual Evaluation}
The next two groups show (\emph{i})~how continuous fine-tuning of \XLMR on multi-choice machine reading comprehension and multi-choice science QA helps, and (\emph{ii})~how the different models (\XLMR, \XLMRb, and mBERT) compare. 
We follow a standard training scheme for such tasks:
first we fine-tune on RACE~\citep{lai-etal-2017-race} ($\sim$85k EN questions over documents), 
then on the AI2 English science datasets (we call them SciENs for shorter), including $\sim$9k EN questions with provided relevant contexts\footnote{We use the data described at {\scriptsize \url{http://leaderboard.allenai.org/arc/submission/blcotvl7rrltlue6bsv0}}}, and, finally, on our multilingual training set (see Section \ref{subsec:datasplits})
with retrieved relevant contexts from Wikipedia (see Appendix \ref{sec:wikicorpus}), which is our desired multilingual evaluation setting and we call it \textit{Full}. 
We can also see that training on the SciENs, which has mostly primary school questions from Natural Sciences,
only yields +0.5\% improvement on \ourdataset.
Nevertheless, we see a 2.4\% improvement with multilingual fine-tuning on \ourdataset and +0.5\% for English.
In the third group, we compare the results from mBERT, \XLMRb, and \XLMR after fine-tuning.  
Increasing the capacity of the model yields improvements: \XLMR scores 7.4\% higher on \ourdataset, and more than 14\% on English datasets, compared to its base version (\XLMRb). However, mBERT and \XLMRb have close performance, with mBERT having a small advantage in the multilingual setting. 

Finally, we fine-tuned mBERT on \ourdataset only. As expected, the performance drops by 3\% absolute compared to the \textit{Full} setup.

\subsection{Knowledge Evaluation}

The last two rows of Table~\ref{tab:res:overall-lang} evaluate the knowledge in the best model, namely \XLMR. 
With \textit{\XLMR as KB} (see Section \ref{subsec:no_training}) we see small
improvement over the random baseline: +5\% ARC Easy, 2\% on R12, and just +1\% on \ourdataset and ARC Challenge. 
Furthermore, we evaluate the knowledge contained in the model after the \textit{Full} fine-tuning by excluding the relevant knowledge context (\textit{ctx}).
This is better than the \textit{\XLMR as KB},
but it still achieves inferior overall results,
which shows that the stored knowledge is not enough, and that we need to explicitly obtain additional knowledge from an external source.

\subsection{Cross-lingual Evaluation}
\label{subsec:cross-lingual-eval}

Table~\ref{tab:res:cross-ling-zeroshot} shows the results from the cross-lingual zero-shot transfer compared to the English-only baseline $en_{all}$, from \XLMR fine-tuned on SciEN. 
The languages are ordered by family, and then alphabetically.
We further fine-tune on a single source language and we test on all other languages using the splits described in Subsection~\ref{subsec:datasplits}. 
The results show that the additional fine-tuning on a single language is mostly positive. 
This is notable when fine-tuning on a language with similar linguistic characteristics to the target language, e.g.,~Balto-Slavic: bg-sr, hr-mk, pl-mk, sr-bg. 

We also see gains when the source language contains more questions from largely represented and harder subjects. 
Examples of such are the experiments showing the positive effects of training on Vietnamese and Macedonian as source languages; they both contain such subjects: Biology, History, Chemistry, Physics, and Geography.

This is an indication that the knowledge from the same or from related subjects in a non-related language is preferred over knowledge from non-related subjects from a related language. For the same reasons, Portuguese and Polish show negative effects of fine-tuning on some of the target languages. 
They contain mostly niche subjects such as Professional, Philosophy, Economics, Geology. 
We see a noticeable drop in accuracy for Portuguese almost everywhere, but it has positive effect on languages that contain similar subjects (Biology, Economics) or are from the same language family such as Spanish and Italian (for Portuguese). 
We see the opposite in the Lithuanian-Polish pair, languages from the same family (but different subjects) have negative, or no effect on each other.
Finally, we analyze the results from language pairs containing parallel examples (the underlined values).
Such pairs show consistent improvement (+5 to +10), which suggests that the model learns to align the parallel knowledge from the source language to the target language. However, we also must note that the effect is strongly dependent on the size of the overlapping sets.

\begin{figure*}[ht!]
    \centering
    \minipage{\textwidth}
    \includegraphics[width=0.95\textwidth]{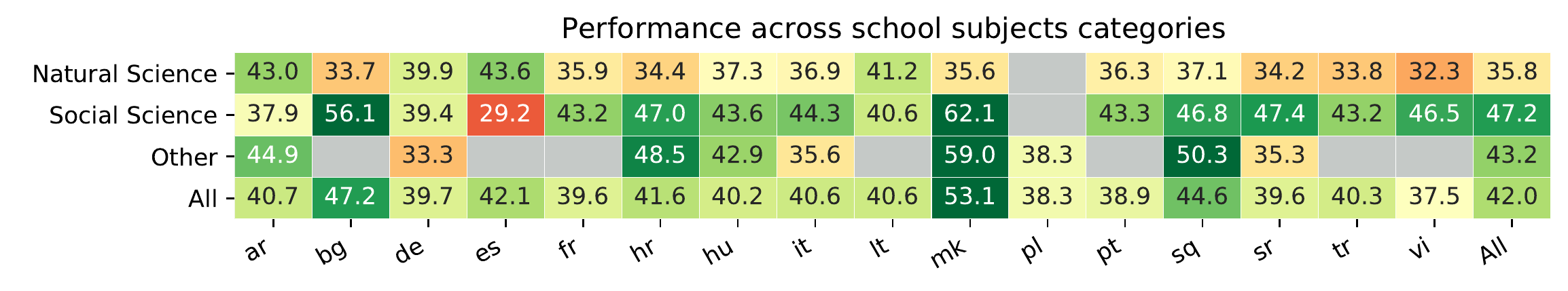}
        
    \endminipage    \hfill
    \minipage{0.90\textwidth}
    \vspace{-2mm}
    \includegraphics[width=\textwidth]{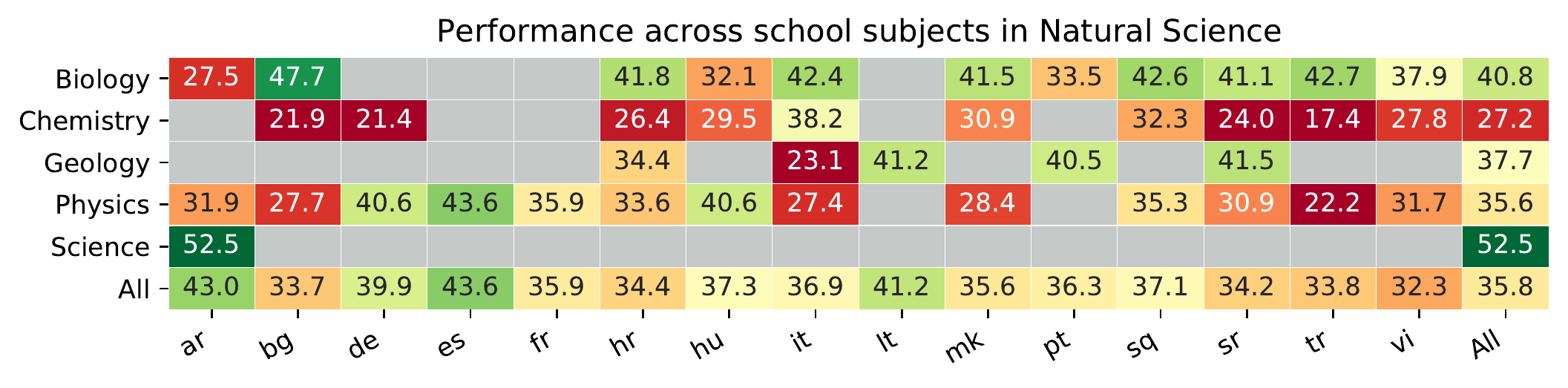}
    \endminipage 
    \vspace{-4mm}
    \caption{Fine-grained evaluation by language and school subjects.}
    \label{fig:heatmap_results}
\end{figure*}

\subsection{Per-subject Fine-grained Evaluation}

Fine-grained evaluation~\citep{mihaylov-frank-2019-discourse, xu-etal-2020-hier-qa} allows an in-depth analysis of the question answering models.
One of the nice features of \ourdataset is that it supports subject-related fine-grained evaluation.
On Figure~\ref{fig:heatmap_results}, the results are shown by subject group and per-subject for Natural 
Science.\footnote{Per-subject results for Social Science and Other are available in Appendix \ref{sec:appendix-fine-grained-eval}.} 

We can see that the Natural Science questions are the most challenging ones, which is mostly due to Chemistry and Physics. 
Those questions require very complex reasoning and knowledge such as understanding physical models, processes and causes, comparisons, algebraic skills and multi-hop reasoning (see Section~\ref{sec:reasoning_types}). 
These skills are currently beyond the capabilities of 
the current QA models, and pose interesting challenges 
for future work~\citep{welbl-etal-2018-constructing,yang-etal-2018-hotpotqa,Saxton2019MathReason,lample2020deep}.
Informatics is another challenging subject, as it requires understanding programming code and positional numerical systems among others.

\section{Discussion}
\label{sec:discussion}

Our results show that initial fine-tuning on a large monolingual out-of-domain multi-choice machine reading comprehension dataset (RACE~\citep{lai-etal-2017-race}) performs much better than \textit{no training} baselines for answering multilingual \ourdataset questions. 
Moreover, additional training on English science QA in lower school levels has no significant effect on the overall accuracy. These results suggest that further investigation of fine-tuning with other multilingual datasets~\citep{Gupta2018MMQAAM,lewis-etal-2020-mlqa,clark2020tydi,efimov2020sberquad,dhoffschmidt2020-fquad,Artetxe:etal:2020-xquad,longpre2020mkqa} is needed in order to understand the domain transfer benefits to science QA in \ourdataset, even if they are not in a multi-choice setting~\citep{khashabi2020unifiedqa}. Using \textit{domain-adaptive} and \textit{task-adaptive pre-training}~\citep{Gururangan2020DontStopPretraining} to the multilingual science QA might offer further potential benefits.

Moreover, we need a better knowledge context for a given question--choice pair (the last row in Table~\ref{tab:res:overall-lang}). Knowing that the context retrieved from the noisy Wikipedia corpus is relevant for answering \ourdataset questions, suggests that we need a better multilingual science corpus, similar to~\citet{Clark2018ThinkYH,pan2019improving,Bhakthavatsalam2020GenericsKBAK}. We further need better multilingual knowledge selection and ranking~\citep{Banerjee2019CarefulSO}. Finally, our cross-lingual experiments show that we can align the knowledge between languages from parallel examples, which poses a new question: \emph{Is it only due to keyword matching or could the model align full sentences?}
\section{Conclusion and Future Work}
\label{sec:concl}

We presented \ourdataset, a new challenging  cross-lingual and multilingual benchmark for science QA in 16 languages and 24 subjects from high school examinations. 

We further proposed new fine-grained evaluation that allows precise comparison across different languages and school subjects. 
We performed various experiments and analysis with pre-trained multilingual models (\XLMR, mBERT), and we demonstrated that there is a need for better reasoning and knowledge transfer in order to solve some of the questions from \ourdataset. 
We hope that our publicly available data and code will enable work on multilingual models that can reason about question answering in the challenging science domain.  

In future work, we plan to extend the dataset with more questions, more subjects, and more languages. We further plan to develop new models to address the specific challenges we identified.

\section*{Acknowledgments}
\label{sec:ack}

We thank the anonymous reviewers for their helpful questions and comments. 
We thank the AI2 Aristo Team for providing the data splits used for pre-training on SciEN datasets. 

This research is partially supported by Project UNITe BG05M2OP001-1.001-0004 funded by the OP ``Science and Education for Smart Growth'' and co-funded by the EU through the ESI Funds.

\bibliography{emnlp2020}

\begin{thebibliography}{60}
\expandafter\ifx\csname natexlab\endcsname\relax\def\natexlab#1{#1}\fi

\bibitem[{Artetxe et~al.(2020)Artetxe, Ruder, and
  Yogatama}]{Artetxe:etal:2020-xquad}
Mikel Artetxe, Sebastian Ruder, and Dani Yogatama. 2020.
\newblock On the cross-lingual transferability of monolingual representations.
\newblock In \emph{Proceedings of the 58th Annual Meeting of the Association
  for Computational Linguistics}, ACL~'20, pages 4623--4637.

\bibitem[{Banerjee et~al.(2019)Banerjee, Pal, Mitra, and
  Baral}]{Banerjee2019CarefulSO}
Pratyay Banerjee, Kuntal~Kumar Pal, Arindam Mitra, and Chitta Baral. 2019.
\newblock Careful selection of knowledge to solve open book question answering.
\newblock In \emph{Proceedings of the 57th Annual Meeting of the Association
  for Computational Linguistics}, ACL~'19, pages 6120--6129, Florence, Italy.

\bibitem[{Bhakthavatsalam et~al.(2020)Bhakthavatsalam, Anastasiades, and
  Clark}]{Bhakthavatsalam2020GenericsKBAK}
Sumithra Bhakthavatsalam, Chloe Anastasiades, and Peter Clark. 2020.
\newblock {G}enerics{KB}: A knowledge base of generic statements.
\newblock \emph{ArXiv}, abs/2005.00660.

\bibitem[{Boratko et~al.(2018)Boratko, Padigela, Mikkilineni, Yuvraj, Das,
  McCallum, Chang, Fokoue, Kapanipathi, Mattei, Musa, Talamadupula, and
  Witbrock}]{Boratko2018-sciq-annotations}
Michael Boratko, Harshit Padigela, Divyendra Mikkilineni, Pritish Yuvraj,
  Rajarshi Das, Andrew McCallum, Maria Chang, Achille Fokoue, Pavan
  Kapanipathi, Nicholas Mattei, Ryan Musa, Kartik Talamadupula, and Michael
  Witbrock. 2018.
\newblock An interface for annotating science questions.
\newblock In \emph{Proceedings of the 2018 Conference on Empirical Methods in
  Natural Language Processing: System Demonstrations}, EMNLP~'18, pages
  102--107, Brussels, Belgium.

\bibitem[{Carrino et~al.(2020)Carrino, Costa-juss{\`a}, and
  Fonollosa}]{carrino2020-spanishquad}
Casimiro~Pio Carrino, Marta~R. Costa-juss{\`a}, and Jos{\'e} A.~R. Fonollosa.
  2020.
\newblock Automatic {S}panish translation of {SQ}u{AD} dataset for
  multi-lingual question answering.
\newblock In \emph{Proceedings of the 12th Language Resources and Evaluation
  Conference}, LREC~'20, pages 5515--5523, Marseille, France.

\bibitem[{Clark et~al.(2020)Clark, Choi, Collins, Garrette, Kwiatkowski,
  Nikolaev, and Palomaki}]{clark2020tydi}
Jonathan~H. Clark, Eunsol Choi, Michael Collins, Dan Garrette, Tom Kwiatkowski,
  Vitaly Nikolaev, and Jennimaria Palomaki. 2020.
\newblock {T}y{D}i {QA}: A benchmark for information-seeking question answering
  in typologically diverse languages.
\newblock \emph{Transactions of the Association for Computational Linguistics},
  8:454--470.

\bibitem[{Clark(2015)}]{clark2015elementary-science}
Peter Clark. 2015.
\newblock Elementary school science and math tests as a driver for {AI}: Take
  the {A}risto challenge!
\newblock In \emph{Proceedings of the Twenty-Ninth Conference on Artificial
  Intelligence}, AAAI~'15, pages 4019--4021, Austin, Texas, USA.

\bibitem[{Clark et~al.(2014)Clark, Balasubramanian, Bhakthavatsalam, Humphreys,
  Kinkead, and Sabharwal}]{Clark2014automatic-kb}
Peter Clark, Niranjan Balasubramanian, Sumithra Bhakthavatsalam, Kevin
  Humphreys, J.~Clint Kinkead, and Ashish Sabharwal. 2014.
\newblock Automatic construction of inference-supporting knowledge bases.
\newblock In \emph{Proceedings of 4th Workshop on Automated Knowledge Base
  Construction}, AKBC~'14, Montreal, Canada.

\bibitem[{Clark et~al.(2018)Clark, Cowhey, Etzioni, Khot, Sabharwal, Schoenick,
  and Tafjord}]{Clark2018ThinkYH}
Peter Clark, Isaac Cowhey, Oren Etzioni, Tushar Khot, Ashish Sabharwal, Carissa
  Schoenick, and Oyvind Tafjord. 2018.
\newblock Think you have solved question answering? {T}ry {ARC}, the {AI2}
  reasoning challenge.
\newblock \emph{ArXiv}, abs/1803.05457.

\bibitem[{Clark et~al.(2019)Clark, Etzioni, Khashabi, Khot, Mishra, Richardson,
  Sabharwal, Schoenick, Tafjord, Tandon, Bhakthavatsalam, Groeneveld, Guerquin,
  and Schmitz}]{Clark2019AristoStory}
Peter Clark, Oren Etzioni, Daniel Khashabi, Tushar Khot, Bhavana~Dalvi Mishra,
  Kyle Richardson, Ashish Sabharwal, Carissa Schoenick, Oyvind Tafjord, Niket
  Tandon, Sumithra Bhakthavatsalam, Dirk Groeneveld, Michal Guerquin, and
  Michael Schmitz. 2019.
\newblock From {`F'} to {`A'} on the {N.Y.} regents science exams: {A}n
  overview of the {A}risto project.
\newblock \emph{ArXiv}, abs/1909.01958.

\bibitem[{Clark et~al.(2016)Clark, Etzioni, Khot, Sabharwal, Tafjord, Turney,
  and Khashabi}]{Clark2016Retrieval}
Peter Clark, Oren Etzioni, Tushar Khot, Ashish Sabharwal, Oyvind Tafjord, Peter
  Turney, and Daniel Khashabi. 2016.
\newblock Combining retrieval, statistics, and inference to answer elementary
  science questions.
\newblock In \emph{Proceedings of the Thirtieth AAAI Conference on Artificial
  Intelligence}, AAAI~’16, page 2580–2586, Phoenix, Arizona, USA.

\bibitem[{Conneau et~al.(2020)Conneau, Khandelwal, Goyal, Chaudhary, Wenzek,
  Guzm{\'a}n, Grave, Ott, Zettlemoyer, and Stoyanov}]{conneau2020-xlm-roberta}
Alexis Conneau, Kartikay Khandelwal, Naman Goyal, Vishrav Chaudhary, Guillaume
  Wenzek, Francisco Guzm{\'a}n, Edouard Grave, Myle Ott, Luke Zettlemoyer, and
  Veselin Stoyanov. 2020.
\newblock Unsupervised cross-lingual representation learning at scale.
\newblock In \emph{Proceedings of the 58th Annual Meeting of the Association
  for Computational Linguistics}, ACL~'20, pages 8440--8451.

\bibitem[{Dalvi et~al.(2019)Dalvi, Tandon, Bosselut, Yih, and
  Clark}]{Mishra2019ProPara}
Bhavana Dalvi, Niket Tandon, Antoine Bosselut, Wen-tau Yih, and Peter Clark.
  2019.
\newblock Everything happens for a reason: Discovering the purpose of actions
  in procedural text.
\newblock In \emph{Proceedings of the 2019 Conference on Empirical Methods in
  Natural Language Processing and the 9th International Joint Conference on
  Natural Language Processing}, EMNLP-IJCNLP~'19, pages 4496--4505, Hong Kong,
  China.

\bibitem[{Devlin et~al.(2019)Devlin, Chang, Lee, and
  Toutanova}]{devlin2019bert}
Jacob Devlin, Ming-Wei Chang, Kenton Lee, and Kristina Toutanova. 2019.
\newblock {BERT}: Pre-training of deep bidirectional transformers for language
  understanding.
\newblock In \emph{Proceedings of the 2019 Conference of the North {A}merican
  Chapter of the Association for Computational Linguistics: Human Language
  Technologies}, NAACL-HLT~'19, pages 4171--4186, Minneapolis, Minnesota, USA.

\bibitem[{d'Hoffschmidt et~al.(2020)d'Hoffschmidt, Vidal, Belblidia, and
  Brendl{\'e}}]{dhoffschmidt2020-fquad}
Martin d'Hoffschmidt, Maxime Vidal, Wacim Belblidia, and Tom Brendl{\'e}. 2020.
\newblock {FQuAD}: {F}rench question answering dataset.
\newblock \emph{ArXiv}, abs/2002.06071.

\bibitem[{Efimov et~al.(2020)Efimov, Chertok, Boytsov, and
  Braslavski}]{efimov2020sberquad}
Pavel Efimov, Andrey Chertok, Leonid Boytsov, and Pavel Braslavski. 2020.
\newblock {SberQuAD} -- {R}ussian reading comprehension dataset: Description
  and analysis.
\newblock In \emph{Proceedings of the 11th International Conference of the CLEF
  Association: Experimental IR Meets Multilinguality, Multimodality, and
  Interaction}, CLEF~'20, pages 3--15, Thessaloniki, Greece.

\bibitem[{Gupta et~al.(2018)Gupta, Kumari, Ekbal, and
  Bhattacharyya}]{Gupta2018MMQAAM}
Deepak Gupta, Surabhi Kumari, Asif Ekbal, and Pushpak Bhattacharyya. 2018.
\newblock {MMQA}: A multi-domain multi-lingual question-answering framework for
  {E}nglish and {H}indi.
\newblock In \emph{Proceedings of the Eleventh International Conference on
  Language Resources and Evaluation}, LREC~'18, pages 2777--2784, Miyazaki,
  Japan.

\bibitem[{Gururangan et~al.(2020)Gururangan, Marasovi{\'c}, Swayamdipta, Lo,
  Beltagy, Downey, and Smith}]{Gururangan2020DontStopPretraining}
Suchin Gururangan, Ana Marasovi{\'c}, Swabha Swayamdipta, Kyle Lo, Iz~Beltagy,
  Doug Downey, and Noah~A. Smith. 2020.
\newblock Don{'}t stop pretraining: Adapt language models to domains and tasks.
\newblock In \emph{Proceedings of the 58th Annual Meeting of the Association
  for Computational Linguistics}, ACL~'20, pages 8342--8360.

\bibitem[{Hardalov et~al.(2019)Hardalov, Koychev, and
  Nakov}]{hardalov-etal-2019-beyond}
Momchil Hardalov, Ivan Koychev, and Preslav Nakov. 2019.
\newblock Beyond {E}nglish-only reading comprehension: Experiments in zero-shot
  multilingual transfer for {B}ulgarian.
\newblock In \emph{Proceedings of the International Conference on Recent
  Advances in Natural Language Processing}, RANLP~'19, pages 447--459, Varna,
  Bulgaria.

\bibitem[{Hu et~al.(2020)Hu, Ruder, Siddhant, Neubig, Firat, and
  Johnson}]{hu2020xtreme}
Junjie Hu, Sebastian Ruder, Aditya Siddhant, Graham Neubig, Orhan Firat, and
  Melvin Johnson. 2020.
\newblock {XTREME}: A massively multilingual multi-task benchmark for
  evaluating cross-lingual generalization.
\newblock In \emph{Proceedings of Machine Learning Research}, ICML~'20, Online.

\bibitem[{Jing et~al.(2019)Jing, Xiong, and Yan}]{jing-etal-2019-bipar}
Yimin Jing, Deyi Xiong, and Zhen Yan. 2019.
\newblock {B}i{P}a{R}: A bilingual parallel dataset for multilingual and
  cross-lingual reading comprehension on novels.
\newblock In \emph{Proceedings of the 2019 Conference on Empirical Methods in
  Natural Language Processing and the 9th International Joint Conference on
  Natural Language Processing}, EMNLP-IJCNLP~'19, pages 2452--2462, Hong Kong,
  China.

\bibitem[{Khashabi et~al.(2016)Khashabi, Khot, Sabharwal, Clark, Etzioni, and
  Roth}]{Khashabi2016QuestionAV}
Daniel Khashabi, Tushar Khot, Ashish Sabharwal, Peter Clark, Oren Etzioni, and
  Dan Roth. 2016.
\newblock Question answering via integer programming over semi-structured
  knowledge.
\newblock In \emph{Proceedings of the Twenty-fifth International Joint
  Conferences on Artificial Intelligence Organization}, IJCAI~'16, pages
  1145--1152, New York, New York.

\bibitem[{Khashabi et~al.(2018)Khashabi, Khot, Sabharwal, and
  Roth}]{khashabi2018question}
Daniel Khashabi, Tushar Khot, Ashish Sabharwal, and Dan Roth. 2018.
\newblock Question answering as global reasoning over semantic abstractions.
\newblock In \emph{Proceedings of the Thirty-Second AAAI Conference on
  Artificial Intelligence}, AAAI~'18, pages 1905--1914, New Orleans, Louisiana,
  USA.

\bibitem[{Khashabi et~al.(2020)Khashabi, Khot, Sabharwal, Tafjord, Clark, and
  Hajishirzi}]{khashabi2020unifiedqa}
Daniel Khashabi, Tushar Khot, Ashish Sabharwal, Oyvind Tafjord, Peter Clark,
  and Hannaneh Hajishirzi. 2020.
\newblock {UnifiedQA}: Crossing format boundaries with a single {QA} system.
\newblock \emph{Finding of EMNLP}.

\bibitem[{Khot et~al.(2020)Khot, Clark, Guerquin, Jansen, and
  Sabharwal}]{Khot2020QASCAD}
Tushar Khot, Peter Clark, Michal Guerquin, Paul~Edward Jansen, and Ashish
  Sabharwal. 2020.
\newblock {QASC}: {A} dataset for question answering via sentence composition.
\newblock In \emph{Proceedings of the Thirty-Fourth AAAI Conference on
  Artificial Intelligence}, AAAI~'20, pages 8082--8090, New York, New York,
  USA.

\bibitem[{Khot et~al.(2017)Khot, Sabharwal, and Clark}]{Khot2017sci-ie}
Tushar Khot, Ashish Sabharwal, and Peter Clark. 2017.
\newblock Answering complex questions using open information extraction.
\newblock In \emph{Proceedings of the 55th Annual Meeting of the Association
  for Computational Linguistics}, ACL~'17, pages 311--316, Vancouver, Canada.

\bibitem[{Khot et~al.(2018)Khot, Sabharwal, and Clark}]{Khot2018SciTaiLAT}
Tushar Khot, Ashish Sabharwal, and Peter Clark. 2018.
\newblock {SciTail}: A textual entailment dataset from science question
  answering.
\newblock In \emph{Proceedings of the Thirty-Second AAAI Conference on
  Artificial Intelligence}, AAAI~'18, pages 5189--5197, New Orleans, Louisiana,
  USA.

\bibitem[{Lai et~al.(2017)Lai, Xie, Liu, Yang, and Hovy}]{lai-etal-2017-race}
Guokun Lai, Qizhe Xie, Hanxiao Liu, Yiming Yang, and Eduard Hovy. 2017.
\newblock {RACE}: Large-scale {R}e{A}ding comprehension dataset from
  examinations.
\newblock In \emph{Proceedings of the 2017 Conference on Empirical Methods in
  Natural Language Processing}, EMNLP~'17, pages 785--794, Copenhagen, Denmark.

\bibitem[{Lample and Charton(2020)}]{lample2020deep}
Guillaume Lample and François Charton. 2020.
\newblock Deep learning for symbolic mathematics.
\newblock In \emph{Proceedings of the 8th International Conference on Learning
  Representations}, ICLR~'20.

\bibitem[{Lan et~al.(2020)Lan, Chen, Goodman, Gimpel, Sharma, and
  Soricut}]{lan2020albert}
Zhenzhong Lan, Mingda Chen, Sebastian Goodman, Kevin Gimpel, Piyush Sharma, and
  Radu Soricut. 2020.
\newblock {ALBERT}: {A} lite {BERT} for self-supervised learning of language
  representations.
\newblock In \emph{Proceedings of the 8th International Conference on Learning
  Representations}, ICLR~'20.

\bibitem[{Lewis et~al.(2020)Lewis, Oguz, Rinott, Riedel, and
  Schwenk}]{lewis-etal-2020-mlqa}
Patrick Lewis, Barlas Oguz, Ruty Rinott, Sebastian Riedel, and Holger Schwenk.
  2020.
\newblock {MLQA}: Evaluating cross-lingual extractive question answering.
\newblock In \emph{Proceedings of the 58th Annual Meeting of the Association
  for Computational Linguistics}, ACL~'20, pages 7315--7330.

\bibitem[{Lim et~al.(2019)Lim, Kim, and Lee}]{lim2019-korquad1}
Seungyoung Lim, Myungji Kim, and Jooyoul Lee. 2019.
\newblock {KorQuAD1.0}: {K}orean {QA} dataset for machine reading
  comprehension.
\newblock \emph{ArXiv}, abs/1909.07005.

\bibitem[{Liu et~al.(2019{\natexlab{a}})Liu, Lin, Liu, and
  Sun}]{liu-etal-2019-xqa}
Jiahua Liu, Yankai Lin, Zhiyuan Liu, and Maosong Sun. 2019{\natexlab{a}}.
\newblock {XQA}: A cross-lingual open-domain question answering dataset.
\newblock In \emph{Proceedings of the 57th Annual Meeting of the Association
  for Computational Linguistics}, ACL~'19, pages 2358--2368, Florence, Italy.

\bibitem[{Liu et~al.(2019{\natexlab{b}})Liu, Deng, Zhu, and Hu}]{liu2019xcmrc}
Pengyuan Liu, Yuning Deng, Chenghao Zhu, and Han Hu. 2019{\natexlab{b}}.
\newblock {XCMRC}: Evaluating cross-lingual machine reading comprehension.
\newblock In \emph{Proceedings of the International Conference on Natural
  Language Processing and Chinese Computing}, NLPCC~'19, pages 552--564,
  Dunhuang, China.

\bibitem[{Liu et~al.(2019{\natexlab{c}})Liu, Ott, Goyal, Du, Joshi, Chen, Levy,
  Lewis, Zettlemoyer, and Stoyanov}]{liu2019roberta}
Yinhan Liu, Myle Ott, Naman Goyal, Jingfei Du, Mandar Joshi, Danqi Chen, Omer
  Levy, Mike Lewis, Luke Zettlemoyer, and Veselin Stoyanov. 2019{\natexlab{c}}.
\newblock {RoBERTa}: {A} robustly optimized {BERT} pretraining approach.
\newblock \emph{ArXiv}, abs/1907.11692.

\bibitem[{Longpre et~al.(2020)Longpre, Lu, and Daiber}]{longpre2020mkqa}
Shayne Longpre, Yi~Lu, and Joachim Daiber. 2020.
\newblock {MKQA}: A linguistically diverse benchmark for multilingual open
  domain question answering.
\newblock \emph{ArXiv}, abs/2007.15207.

\bibitem[{Mihaylov et~al.(2018)Mihaylov, Clark, Khot, and
  Sabharwal}]{mihaylov-etal-2018-suit}
Todor Mihaylov, Peter Clark, Tushar Khot, and Ashish Sabharwal. 2018.
\newblock Can a suit of armor conduct electricity? {A} new dataset for open
  book question answering.
\newblock In \emph{Proceedings of the Conference on Empirical Methods in
  Natural Language Processing}, EMNLP~'18, pages 2381--2391, Brussels, Belgium.

\bibitem[{Mihaylov and Frank(2019)}]{mihaylov-frank-2019-discourse}
Todor Mihaylov and Anette Frank. 2019.
\newblock Discourse-aware semantic self-attention for narrative reading
  comprehension.
\newblock In \emph{Proceedings of the 2019 Conference on Empirical Methods in
  Natural Language Processing and the 9th International Joint Conference on
  Natural Language Processing}, EMNLP~'19, pages 2541--2552, Hong Kong, China.

\bibitem[{Mitra et~al.(2019)Mitra, Clark, Tafjord, and
  Baral}]{Mitra2019DeclarativeQA}
Arindam Mitra, Peter Clark, Oyvind Tafjord, and Chitta Baral. 2019.
\newblock Declarative question answering over knowledge bases containing
  natural language text with answer set programming.
\newblock In \emph{Proceedings of the Thirty-Third AAAI Conference on
  Artificial Intelligence}, AAAI~'19, pages 3003--3010, Honolulu, Hawaii, USA.

\bibitem[{Ni et~al.(2019)Ni, Zhu, Chen, and McAuley}]{ni-etal-2019-learning}
Jianmo Ni, Chenguang Zhu, Weizhu Chen, and Julian McAuley. 2019.
\newblock Learning to attend on essential terms: An enhanced retriever-reader
  model for open-domain question answering.
\newblock In \emph{Proceedings of the Conference of the North {A}merican
  Chapter of the Association for Computational Linguistics: Human Language
  Technologies}, NAACL-HLT~'19, pages 335--344, Minneapolis, Minnesota, USA.

\bibitem[{Pan et~al.(2019)Pan, Sun, Yu, Chen, Ji, Cardie, and
  Yu}]{pan2019improving}
Xiaoman Pan, Kai Sun, Dian Yu, Jianshu Chen, Heng Ji, Claire Cardie, and Dong
  Yu. 2019.
\newblock Improving question answering with external knowledge.
\newblock In \emph{Proceedings of the 2nd Workshop on Machine Reading for
  Question Answering}, MRQA~'19, pages 27--37, Hong Kong, China.

\bibitem[{Peters et~al.(2018)Peters, Neumann, Iyyer, Gardner, Clark, Lee, and
  Zettlemoyer}]{Peters:2018:ELMo}
Matthew Peters, Mark Neumann, Mohit Iyyer, Matt Gardner, Christopher Clark,
  Kenton Lee, and Luke Zettlemoyer. 2018.
\newblock Deep contextualized word representations.
\newblock In \emph{Proceedings of the 2018 Conference of the North American
  Chapter of the Association for Computational Linguistics: Human Language
  Technologies}, NAACL-HLT~'18, pages 2227--2237, New Orleans, Louisiana, USA.

\bibitem[{Petroni et~al.(2019)Petroni, Rockt{\"a}schel, Riedel, Lewis, Bakhtin,
  Wu, and Miller}]{petroni2019language}
Fabio Petroni, Tim Rockt{\"a}schel, Sebastian Riedel, Patrick Lewis, Anton
  Bakhtin, Yuxiang Wu, and Alexander Miller. 2019.
\newblock Language models as knowledge bases?
\newblock In \emph{Proceedings of the 2019 Conference on Empirical Methods in
  Natural Language Processing and the 9th International Joint Conference on
  Natural Language Processing}, EMNLP-IJCNLP~'19, pages 2463--2473, Hong Kong,
  China.

\bibitem[{Radford et~al.(2018)Radford, Narasimhan, Salimans, and
  Sutskever}]{radford2018gpt1}
Alec Radford, Karthik Narasimhan, Tim Salimans, and Ilya Sutskever. 2018.
\newblock Improving language understanding by generative pre-training.

\bibitem[{Radford et~al.(2019)Radford, Wu, Child, Luan, Amodei, and
  Sutskever}]{radford2019language}
Alec Radford, Jeffrey Wu, Rewon Child, David Luan, Dario Amodei, and Ilya
  Sutskever. 2019.
\newblock Language models are unsupervised multitask learners.
\newblock \emph{OpenAI Blog}.

\bibitem[{Raffel et~al.(2020)Raffel, Shazeer, Roberts, Lee, Narang, Matena,
  Zhou, Li, and Liu}]{Raffel2020ExploringTTT_T5}
Colin Raffel, Noam Shazeer, Adam Roberts, Katherine Lee, Sharan Narang, Michael
  Matena, Yanqi Zhou, Wei Li, and Peter~J. Liu. 2020.
\newblock Exploring the limits of transfer learning with a unified text-to-text
  transformer.
\newblock \emph{Journal of Machine Learning Research}, 21(140):1--67.

\bibitem[{Rajpurkar et~al.(2016)Rajpurkar, Zhang, Lopyrev, and
  Liang}]{Rajpurkar2016SQuAD10}
Pranav Rajpurkar, Jian Zhang, Konstantin Lopyrev, and Percy Liang. 2016.
\newblock {SQ}u{AD}: 100,000+ questions for machine comprehension of text.
\newblock In \emph{Proceedings of the 2016 Conference on Empirical Methods in
  Natural Language Processing}, EMNLP~'16, pages 2383--2392, Austin, Texas,
  USA.

\bibitem[{Rogers et~al.(2020)Rogers, Kovaleva, and
  Rumshisky}]{rogers2020primer}
Anna Rogers, Olga Kovaleva, and Anna Rumshisky. 2020.
\newblock A primer in {BERTology}: What we know about how {BERT} works.
\newblock \emph{ArXiv}, abs/2002.12327.

\bibitem[{Saxton et~al.(2019)Saxton, Grefenstette, Hill, and
  Kohli}]{Saxton2019MathReason}
David Saxton, Edward Grefenstette, Felix Hill, and Pushmeet Kohli. 2019.
\newblock Analysing mathematical reasoning abilities of neural models.
\newblock In \emph{Proceedings of the 7th International Conference on Learning
  Representations}, ICLR~'19, New Orleans, Louisiana, USA.

\bibitem[{Schoenick et~al.(2017)Schoenick, Clark, Tafjord, Turney, and
  Etzioni}]{Schoenick2017ai2-turing}
Carissa Schoenick, Peter Clark, Oyvind Tafjord, Peter~D. Turney, and Oren
  Etzioni. 2017.
\newblock Moving beyond the {T}uring test with the {A}llen {AI} {S}cience
  {C}hallenge.
\newblock \emph{Communications of the ACM}, 60:60 -- 64.

\bibitem[{Sun et~al.(2019)Sun, Yu, Yu, and Cardie}]{sun2019readingstrategies}
Kai Sun, Dian Yu, Dong Yu, and Claire Cardie. 2019.
\newblock Improving machine reading comprehension with general reading
  strategies.
\newblock In \emph{Proceedings of the 2019 Conference of the North {A}merican
  Chapter of the Association for Computational Linguistics: Human Language
  Technologies}, NAACL-HLT~'19, pages 2633--2643, Minneapolis, Minnesota, USA.

\bibitem[{Tafjord et~al.(2019)Tafjord, Clark, Gardner, tau Yih, and
  Sabharwal}]{Tafjord2019QuaRelAD}
Oyvind Tafjord, Peter Clark, Matt Gardner, Wen tau Yih, and Ashish Sabharwal.
  2019.
\newblock {Q}ua{R}el: {A} dataset and models for answering questions about
  qualitative relationships.
\newblock In \emph{Proceedings of the Thirty-Third AAAI Conference on
  Artificial Intelligence}, AAAI~'19, pages 7064--7071, Honolulu, Hawaii, USA.

\bibitem[{Van~Nguyena et~al.(2020)Van~Nguyena, Trana, Luua, and
  Gia-Tuan}]{nguyen2020viet-qa-dataset}
Kiet Van~Nguyena, Khiem~Vinh Trana, Son~T Luua, and Anh Gia-Tuan. 2020.
\newblock Enhancing lexical-based approach with external knowledge for
  {V}ietnamese multiple-choice reading comprehension.
\newblock \emph{ArXiv}, abs/2001.05687.

\bibitem[{Vaswani et~al.(2017)Vaswani, Shazeer, Parmar, Uszkoreit, Jones,
  Gomez, Kaiser, and Polosukhin}]{NIPS2017_7181:transformer}
Ashish Vaswani, Noam Shazeer, Niki Parmar, Jakob Uszkoreit, Llion Jones,
  Aidan~N Gomez, {\L}ukasz Kaiser, and Illia Polosukhin. 2017.
\newblock Attention is all you need.
\newblock In \emph{Proceedings of the Annual Conference on Neural Information
  Processing Systems}, NIPS~'17, pages 5998--6008, Long Beach, California, USA.

\bibitem[{Welbl et~al.(2017)Welbl, Liu, and Gardner}]{Welbl2017SciQ}
Johannes Welbl, Nelson~F. Liu, and Matt Gardner. 2017.
\newblock Crowdsourcing multiple choice science questions.
\newblock In \emph{Proceedings of the 3rd Workshop on Noisy User-generated
  Text}, W-NUT~'17, pages 94--106, Copenhagen, Denmark.

\bibitem[{Welbl et~al.(2018)Welbl, Stenetorp, and
  Riedel}]{welbl-etal-2018-constructing}
Johannes Welbl, Pontus Stenetorp, and Sebastian Riedel. 2018.
\newblock Constructing datasets for multi-hop reading comprehension across
  documents.
\newblock \emph{Transactions of the Association for Computational Linguistics},
  6:287--302.

\bibitem[{Wolf et~al.(2019)Wolf, Debut, Sanh, Chaumond, Delangue, Moi, Cistac,
  Rault, Louf, Funtowicz, and Brew}]{Wolf2019HuggingFaces-Transformers}
Thomas Wolf, Lysandre Debut, Victor Sanh, Julien Chaumond, Clement Delangue,
  Anthony Moi, Pierric Cistac, Tim Rault, R\'{e}mi Louf, Morgan Funtowicz, and
  Jamie Brew. 2019.
\newblock {HuggingFace}'s {T}ransformers: State-of-the-art natural language
  processing.
\newblock \emph{ArXiv}, abs/1910.03771.

\bibitem[{Xu et~al.(2020)Xu, Jansen, Martin, Xie, Yadav, Tayyar~Madabushi,
  Tafjord, and Clark}]{xu-etal-2020-hier-qa}
Dongfang Xu, Peter Jansen, Jaycie Martin, Zhengnan Xie, Vikas Yadav, Harish
  Tayyar~Madabushi, Oyvind Tafjord, and Peter Clark. 2020.
\newblock Multi-class hierarchical question classification for multiple choice
  science exams.
\newblock In \emph{Proceedings of the 12th Language Resources and Evaluation
  Conference}, LREC~'20, pages 5370--5382, Marseille, France.

\bibitem[{Yang et~al.(2019)Yang, Dai, Yang, Carbonell, Salakhutdinov, and
  Le}]{yang2019xlnet}
Zhilin Yang, Zihang Dai, Yiming Yang, Jaime Carbonell, Russ~R Salakhutdinov,
  and Quoc~V Le. 2019.
\newblock {XLNet}: Generalized autoregressive pretraining for language
  understanding.
\newblock In \emph{Advances in Neural Information Processing Systems 32},
  NIPS~'19, pages 5753--5763. Vancouver, Canada.

\bibitem[{Yang et~al.(2018)Yang, Qi, Zhang, Bengio, Cohen, Salakhutdinov, and
  Manning}]{yang-etal-2018-hotpotqa}
Zhilin Yang, Peng Qi, Saizheng Zhang, Yoshua Bengio, William Cohen, Ruslan
  Salakhutdinov, and Christopher~D. Manning. 2018.
\newblock {H}otpot{QA}: A dataset for diverse, explainable multi-hop question
  answering.
\newblock In \emph{Proceedings of the 2018 Conference on Empirical Methods in
  Natural Language Processing}, EMNLP~'18, pages 2369--2380, Brussels, Belgium.

\end{thebibliography}
\bibliographystyle{acl_natbib}

\clearpage
\appendix

\section{Fine-Tuning and Hyper-parameters}
\label{sec:training-and-hyper-params}

In this work, we are interested in the cross-lingual transferability of multilingual models such as mBERT~\citep{devlin2019bert} and XLM-RoBERTa~\citep{conneau2020-xlm-roberta}, each of which comes pre-trained on more than 100 languages. We evaluated the QA capabilities of these models, following the established protocol~\citep{devlin2019bert, liu2019roberta, sun2019readingstrategies}, namely we fine-tuned them to predict the correct answer in a multi-choice setting, given a selected context. The aforementioned setup feeds the pre-trained model with a text, processed using the model's tokenizer in the following format: 
\begin{quote}
[CLS] C [SEP] Q + O [SEP]
\end{quote}
where C, Q and O are the tokenized \textit{knowledge \textbf{C}ontext} (see Appendix~\ref{sec:wikicorpus}),
\textit{\textbf{Q}uestion}, and \textit{\textbf{O}ption}, respectively.

We used the Transformers library~\citep{Wolf2019HuggingFaces-Transformers}. 
We fine-tuned mBERT, \XLMR, and \XLMRb in three steps. We first fine-tuned the models with RACE~\citep{lai-etal-2017-race}, a multiple-choice reading comprehension dataset with around 85k questions for training. Then, we trained on the combination of ARC~\citep{Clark2018ThinkYH}, OpenBookQA~\citep{mihaylov-etal-2018-suit}, and Regents Living Environments, as in the \textit{AristoRoBERTaV7} ARC Challenge leaderboard entry\footnote{ \url{https://leaderboard.allenai.org/arc/submission/blcotvl7rrltlue6bsv0}}; we refer to these datasets as \textit{SciENs} (\textbf{Sci}ence \textbf{En}glish dataset\textbf{s}).
We used the resulting pre-trained models as base models for our \textit{Multilingual} and \textit{Cross-lingual} evaluations (Section \ref{sec:experiments_and_results} in the paper). 
For the multilingual evaluation, we continued training the model, previously fine-tuned on the SciENs datasets, with our multilingual Train$_{Mul}$ set, validating on Dev$_{Mul}$ and testing on Test$_{Mul}$.
For our cross-lingual evaluation, we continued training the SciENs model on separate languages, as described in Section~\ref{subsec:cross-lingual-eval}.

In Table~\ref{tab:fine-tune-hyper-params}, we show the values of the hyper-parameters for each fine-tuning step and corresponding model. 
Note that these hyper-parameters were \textbf{not} obtained with an exhaustive search, and thus a better setting might exist for each model and dataset. 
Initially, we used the hyper-parameters for \textit{AristoRoBERTaV7} ARC Leaderboard submission for English-only RoBERTa~\citep{liu2019roberta}: epochs = 4, learning rate = 1e-5. 

With these parameters alone, the models did not perform well, and thus we added a warmup of 0.1 and a weight decay of 0.06, which stabilized the training. In all experiments, we used the Adam optimizer with $\beta_1$=0.9, $\beta_2$=0.999, and $\epsilon$=1e-08.

We further performed manual tuning of the hyper-parameters: we experimented with variations thereof, depending on the performance on the corresponding development sets, and we ended up with the values in Table~\ref{tab:fine-tune-hyper-params}. Moreover, we adjusted the batch size and the accumulation steps depending on the availability of the GPUs on our cluster: Nvidia GTX 1080 Ti (Pascal, 11GB memory) or Nvidia Quadro RTX 6000 (24GB).
For each examined setting, we trained for up-to 6 epochs, evaluating the model on the corresponding development set every 100 to 1000 update steps, depending on the dataset size and the effective batch size. For the final evaluations, we chose the model with the highest accuracy score on the corresponding development set. 

Fine-tuning \XLMR (550M parameters) on Nvidia Quadro RTX 6000 (24GB) with the given hyper-parameters took around three hours per epoch when fine-tuned on RACE ($\sim$85k examples), 30 minutes per epoch when fine-tuned on SciENs ($\sim$9k examples), and 30 minutes on \ourdataset on Train$_{Mul}$ ($\sim$8k examples). 
Fine-tuning \XLMRb (270M parameters) and mBERT (172M parameters) on Nvidia GTX 1080 Ti (Pascal, 11GB memory) with the given hyper-parameters took roughly 2 to 2.5 hours per epoch when fine-tuned on RACE ($\sim$85k examples), 30 to 35 minutes per epoch when fine-tuned on SciENs ($\sim$9k examples), and additional 30 minutes on the \ourdataset Train$_{Mul}$ ($\sim$8k examples).

\begin{table*}[t!]
\centering
\small
\begin{tabular}{lcccccc}
\toprule
\bf Model & \bf Batch Size & \bf Accum. Steps & \bf Max Seq. Len. & \bf Learn Rate & \bf Warmup & \bf Weight Decay  \\
\midrule
\multicolumn{6}{c}{fine-tune on RACE (Step 1)}  \\\midrule
mBERT & 4 & 64 & 320 & 0.00005 & 0.1 & -  \\\midrule
\begin{tabular}[c]{@{}l@{}}\XLMR\\ \XLMRb\end{tabular} & 2 & 16 & 320 & 0.00001 & 0.1 & 0.06  \\\midrule
\multicolumn{6}{c}{fine-tune on SciENs (Step 2)} &  \\\midrule
\begin{tabular}[c]{@{}l@{}}mBERT\\ \XLMR\\ \XLMRb\end{tabular} & 2 & 16 & 320 & 0.00001 & 0.2 & 0.06  \\\midrule
\multicolumn{6}{c}{\ourdataset Train$_{Mul}$ (Step 3 - Multilingual)} &  \\\midrule
\begin{tabular}[c]{@{}l@{}}mBERT\\ \XLMR\\ \XLMRb\end{tabular} & 2 & 16 & 320 & 0.00001 & 0.2 & 0.06  \\\midrule
\multicolumn{6}{c}{for each source language  (Step 3 - Cross-lingual)} &  \\\midrule
\begin{tabular}[c]{@{}l@{}}mBERT\\ \XLMR\\ \XLMRb\end{tabular} & 2 & 8 & 320 & 0.00001 & 0.2 & 0.06 \\\bottomrule
\end{tabular}
\caption{The hyper-parameter values we used for fine-tuning.}
\label{tab:fine-tune-hyper-params}
\end{table*}

\section{Subject Analysis}
\label{sec:subject_analysis}

The \emph{Natural Science} group contains five subjects. The corresponding question length is 16.4 characters and 3.9 answers on average. Some of the subjects are well-known and widely studied, such as \textit{Physics, Biology} and \textit{Chemistry}. They appear in at least 10 out of the 16 languages, covering 7 out of 8 language families. However, \textit{Geology} is less common and is present for only 4 languages. Finally, \textit{Science} is an isolated subject for Arabic. This group contributes a total of 9,962 questions in the entire dataset, as shown in Table~\ref{tab:subject_lang_dist}. The major groups in the table are divided with a horizontal line for convenience.

The second subject group covers \emph{Social Sciences}. \textit{Geography, History, Philosophy, Psychology} and \textit{Ethics} are more common, and thus are included in seven languages on average (see Table~\ref{tab:subject_lang_dist}). The subject group's average question length is 18.5 characters. The only sizeable deviation being for Citizenship, as most of the questions in this subject explain some social situation in detail.

The last and smallest of the three subject groups is \emph{Others}. It combines subjects that cannot be categorized as exactly science-related (either social or natural). Those subjects are often specific for a particular country or culture and are fairly diverse. As expected, they are present for less languages (just two).

\begin{table*}[t!]
    \centering
    \small
    \setlength{\tabcolsep}{3pt} % Default 
    \resizebox{\textwidth}{!}{%
    \begin{tabular}{llllrrrrrr}
    \toprule
    \bf Group & \bf Subject & \bf Language & \bf Grade & \bf Q Len & \bf Ch Len & \bf \#Ch & \bf \#Q & \bf Vocab \\
    \midrule
    Natural Science & Biology & ar, bg, hr, hu, it, sr, & H & 18.2 & 4.6 & 4.0 & 3,042 & 24,603 \\
    & & sq, mk, tr, pt, vi & & & & & \\
    Natural Science & Chemistry & bg, hr, it, sr, de, hu, & H & 17.3 & 4.6 & 4.2 & 2,315 & 14,420 \\
    & & sq, mk, tr, vi & & & & & \\
    Natural Science & Geology & hr, it, sr, lt, pt & H & 12.9 & 5.6 & 4.0 & 720 & 7,251 \\
    Natural Science & Physics & ar, bg, hr, it, sr, fr, de, & H & 24.9 & 7.0 & 3.6 & 3,465 & 26,103 \\
    & & hu, es, sq, mk, tr, vi & & & & & \\
    
    Natural Science & Science & ar & M, H & 9.1 & 3.0 & 4.0 & 120 & 1,239 \\
    \midrule
    Social Science & Busin. \& Econ. & fr, de, hu, sq, mk, tr, pt & H & 5.7 & 6.5 & 3.9 & 2,012 & 16,875 \\
    Social Science & Citizenship & vi & H & 45.1 & 6.3 & 4.0 & 119 & 980 \\
    Social Science & Ethics & hr, it, sr & H & 15.5 & 2.6 & 4.0 & 194 & 1,859 \\
    Social Science & Geography & bg, hr, fr, de, hu, it, & H & 15.2 & 5.0 & 4.2 & 1,349 & 11,207 \\
    & & sr, es, tr, vi & & & & & \\
    Social Science & History & bg, hr, it, sr, lt, sq, & H & 16.6 & 5.9 & 4.1 & 3,300 & 32,709 \\
    & & mk, tr, vi & & & & & \\
    Social Science & Philosophy & bg, hr, it, sr, sq, mk, & H & 16.5 & 3.9 & 4.1 & 1,903 & 19,373 \\
    & & tr, pt & & & & & \\
    Social Science & Politics & hr, hu, it, sr & H & 18.2 & 2.8 & 3.0 & 493 & 5,068 \\
    Social Science & Psychology & hr, it, sr & H & 16.5 & 3.9 & 4.1 & 1,903 & 19,373 \\
    Social Science & Social & ar & M, H & 10.8 & 3.4 & 4.0 & 277 & 2,828 \\
    Social Science & Sociology & hr, it, sr, sq, mk, tr & H & 15.2 & 3.4 & 4.0 & 566 & 6,374 \\
    \midrule
    Other & Agriculture & hu & H & 7.9 & 3.6 & 4.3 & 215 & 1,918 \\
    Other & Fine Arts & sq, mk & H & 12.1 & 3.8 & 4.0 & 757 & 5,691 \\
    Other & Forestry & hu & H & 7.8 & 2.9 & 3.7 & 241 & 1,957 \\
    Other & Informatics & hr, it, sr & H & 18.7 & 6.2 & 4.0 & 311 & 2,695 \\
    Other & Islamic Studies & ar & M, H & 9.4 & 3.0 & 4.0 & 78 & 925 \\
    Other & Landscaping & hu & H & 7.4 & 3.8 & 4.9 & 49 & 596 \\
    Other & Professional & pl & H & 13.7 & 4.3 & 4.0 & 1,971 & 18,990 \\
    Other & Religion & hr, sr & H & 10.3 & 3.6 & 4.0 & 222 & 2,159 \\
    Other & Tourism & de, hu & H & 8.8 & 5.2 & 4.0 & 20 & 359 \\
    \bottomrule
    \end{tabular}
    }
    \caption{Per-subject statistics. The grade is High (H), and Middle (M). The average length of the question (\textit{Q Len}) and the choices (\textit{Ch Len}) are measured in number of tokens, and the vocabulary size (\textit{Vocab}) is shown in number of words.}
    \label{tab:subject_lang_dist}
\end{table*}

\subsection{Subject Definitions}
\label{appendix:subject:definitions}

Next, we give a brief description of the less commonly known subjects included in our dataset.

\paragraph{Agriculture} covers questions about soil farming and preservation, small animals breeding and their general health care, and vehicle maintenance and repair. 

\paragraph{Business \& Economics} is a term used to combine five similar subjects related to business and economics. The questions in these subjects cover theoretical questions on economics basis, marketing questions, business questions with elements of accountancy, finances, and organizational studies.
\hfill

\paragraph{Citizenship} is a specific subject from the Vietnamese school system, which tries to inform and give better perspective on different social situations, to educate students in how to perform better, and to be a more aware member of the society by analyzing different norms and personal morality.

\paragraph{Fine Arts} contains analytical and historical questions about different forms of art such as movies, music, art, etc.

\paragraph{Forestry} studies the craft of managing, using, conserving, and repairing forests, woodlands, and associated resources around them such as water sources and soil. 

\paragraph{Geology} is the study of the Earth, with the general exclusion of present-day life, flow within the ocean, and the atmosphere. Questions from this subject cover branches of Geology such as Economical Geology, Marine Geology, Geomorphology, and Geophysics.

\paragraph{Informatics} consists of questions about basic hardware knowledge and software management as well as basics of different positional numeral systems (e.g., binary and hexadecimal).

\paragraph{Islamic Studies} refers to the academic studies of Islam, Quran excerpts, and Muslim morality. This a subject studied in the Qatari educational system during both middle and high school.

\hfill

\paragraph{Landscaping} teaches about modifying the visible features of an area of land, trees and park decorations. It also contains questions about plants and soils.

\paragraph{Politics} covers Croatia's political system, historical questions about the country's development, as well as different regulations and laws, international relations and contracts.

\paragraph{Professional} subject is present in the Polish school system and covers knowledge on specific professions such as flight attendant, babysitter, care taker, office worker in terms of profession's regulations, rules and established norms, etc.

\paragraph{Religion} subject covers Christianity studies such as Bible knowledge, related traditions, e.g.,~baptism, marriage, etc.

\paragraph{Tourism} covers hospitality management, as well as basis of business and traditions in Hungary and its neighbouring countries.

\paragraph{Science} which is used in the Arabic school system throughout middle and high grade studies combines general science questions from Biology, Chemistry, Physics Geology and their branches such as as Biophysics, Astrophysics, and Biochemistry.

\paragraph{Social} subject, similarly to Science, combines questions from political, cultural, historical and geographical studies.

\paragraph{Sociology} is the study of society, patterns of social relationships, social interaction, and culture that surrounds our everyday life.

\begin{table*}[t!]
    \centering
    \small
    \setlength{\tabcolsep}{3pt}
    \begin{tabular}{ll|rrcccc}
    \toprule
        \bf Language & \bf Wiki code  & \bf \#Sentences & \bf \#Articles & \bf Stop word  & \bf Stemming & \bf Keyword & \bf Language \\
        {} &   &  \bf (millions) &  \bf (millions) &   \bf removal &  &  \bf extraction & \bf specific \\
        \midrule
        ARC Corpus & - & 14.6 & - & \Checkmark & \Checkmark & \Checkmark & \Checkmark \\
        \midrule
        German & de & 50.0 & 2.43 &  \Checkmark & \Checkmark & \Checkmark & \Checkmark \\
        French & fr & 30.0 & 2.22 & \Checkmark & \Checkmark & \Checkmark & \Checkmark \\
        Italian & it& 17.5 & 1.61 &  \Checkmark & \Checkmark & \Checkmark & \Checkmark  \\
        Spanish & es & 22.7 & 1.60 &  \Checkmark & \Checkmark & \Checkmark &  \\
        Polish & pl & 15.6 & 1.41 &  & \Checkmark &  &  \\
        Vietnamese & vi & 6.4 & 1.25 & \Checkmark & \Checkmark &  & \Checkmark  \\
        Portuguese & pt & 11.6 & 1.03 &  \Checkmark & \Checkmark & \Checkmark &  \\
        Arabic & ar & 6.0 & 1.04 & \Checkmark & \Checkmark & \Checkmark & \Checkmark \\
        \midrule 
        Serbian & sr & 4.6 & 0.63 &  &  &  & \\
        Hungarian & hu & 7.1 & 0.47 &  \Checkmark & \Checkmark & \Checkmark &   \\
        Turkish & tr & 4.0 & 0.35 &  \Checkmark & \Checkmark & \Checkmark & \Checkmark \\
        Bulgarian & bg & 3.0 & 0.26 &  \Checkmark & \Checkmark & \Checkmark &   \\
        Croatian & hr & 2.7 & 0.22 &  &  &  &  \\
        Lithuanian & lt & 2.0 & 0.20 & \Checkmark & \Checkmark & \Checkmark &   \\
        Macedonian & mk & 1.6 & 0.11 &  &  &  &  \\
        \midrule
        Albanian & sq & 0.8 & 0.08 &  &  &  &  \\
        \bottomrule
    \end{tabular}
    \caption{Description of the per-language indices used as a source of background knowledge in our experiments.}
    \label{tab:wikistats}
\end{table*}
\newpage 

\section{Reasoning and Knowledge Types}
\label{appendix:reasoning_types}

For our reasoning and knowledge type annotations, we followed the procedure and re-used the annotation types presented in \cite{Clark2018ThinkYH,Boratko2018-sciq-annotations}. However, as they were designed mainly for Natural Science questions, we had to extend them with two new
% annotation 
types:

\paragraph{Domain Facts and Knowledge} (Knowledge) This skill requires specific expertise in properties and facts in a given domain, e.g.,~physical properties, characteristics of a chemical element.

\noindent Example from Philosophy (\emph{Portugal}):

\vspace{3mm}
\begin{minipage}{0.44\textwidth}
\small
\emph{Which of the following is an example of a priori knowledge?}\\
A) I know my name. \\
B) I know how old I am. \\
C) \textit{I know that no brother is an only child}. \checkmark \\
D) I know some parents are not married. \\
\end{minipage}

\paragraph{Negation} (Reasoning) is a direct statement of negation, and it is often combined with other reasoning types such as linguistic matching.

\noindent Example from Fine Arts (\emph{North Macedonia}):

\vspace{3mm}
\begin{minipage}{0.44\textwidth}
\small
\emph{Which of the following works of art does \textbf{\textit{not}} belong to the fine arts?}\\
A) Graphics. \\
B) \textit{Poem}. \checkmark \\
C) Design. \\
D) Sculpture. \\
\end{minipage}

\section{Background Knowledge Corpus}
\label{sec:wikicorpus}

Students need good textbooks to study before they can pass an exam, and the same holds for a good machine reading model. However, finding the information needed to answer a question, especially for questions in such a narrow domain as the subjects studied in high schools, usually requires a collection of specialized texts. 
The ARC Corpus~\citep{Clark2018ThinkYH} is an example of such a collection. It is built by querying a major search engine, and around 100 hand-written templates for 80 science topics covered by US elementary and middle schools. Albeit effective, this strategy relies on crafting templates for all language--subject pairs, making the task time-consuming if applied to multiple languages and subjects.

In our work, we used articles from Wikipedia to build a background knowledge corpus for each language. In particular, we parsed the text from the entire Wikipage, removing non-textual content, e.g.,~HTML tags, tables, etc. Following the common strategy used to solve similar tasks in English~\citep{Clark2018ThinkYH,mihaylov-etal-2018-suit}, 
we split each document into sentences and we indexed them using an inverted index. In order to reduce the search space, and to mitigate the effect of known linguistic phenomena within the same language family, e.g., homonyms, partially shared alphabet, etc., we created a separate index for each language.

\begin{figure*}[t!]
    \centering
    \includegraphics[width=\textwidth]{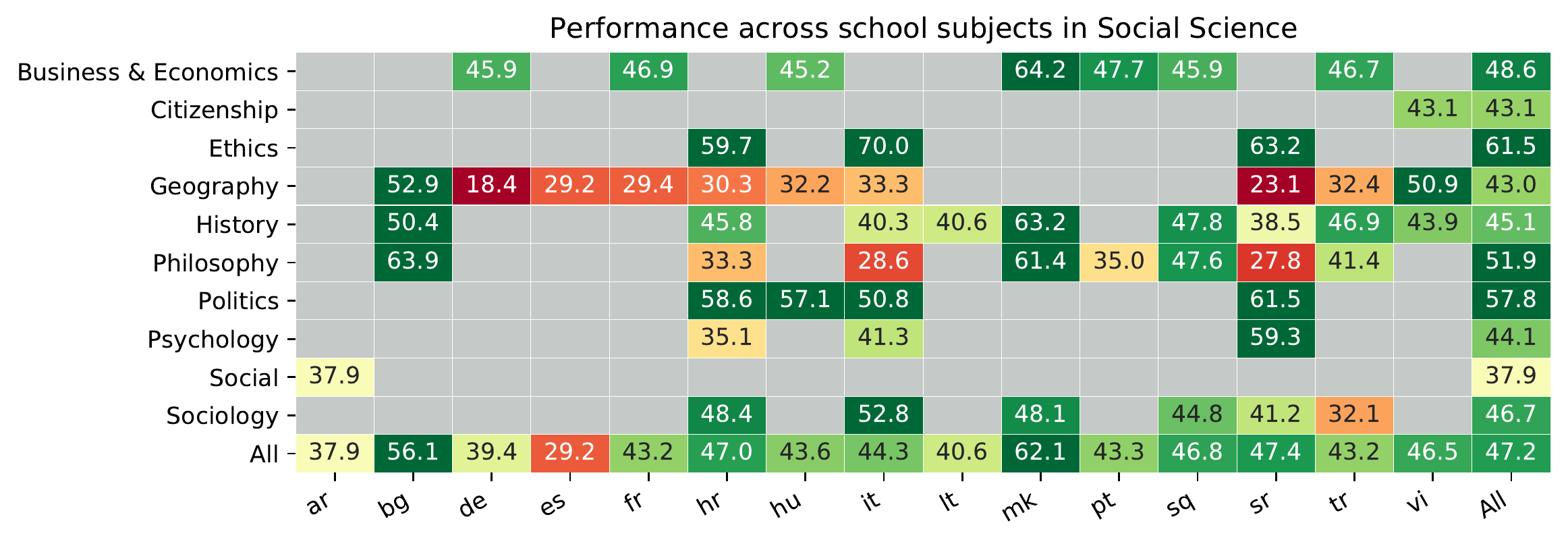}
    \includegraphics[width=0.75\textwidth]{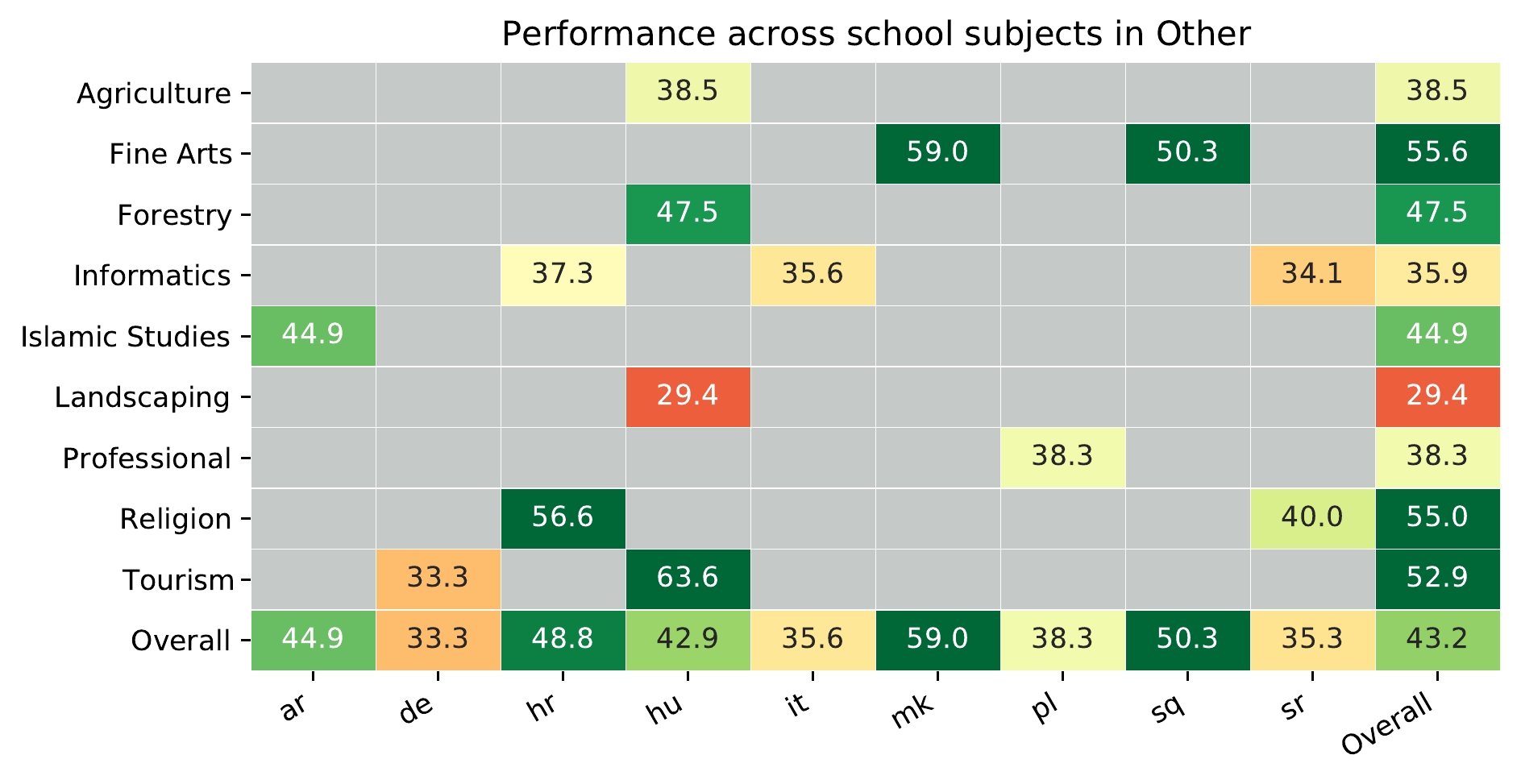}
    \caption{Fine-grained evaluation by language and school subjects in \emph{Social Science} and \emph{Other}.}
    \label{fig:finegrained-othersocial}
\end{figure*}

Table~\ref{tab:wikistats} describes the main characteristics of the indices created for each language from its Wikipedia dump.\footnote{We used the official Wiki dumps from March 2020 for all languages. \url{http://dumps.wikimedia.org/}}
We compared the size of our index to the one from ARC~\citep{Clark2018ThinkYH}. The number of articles for each language is taken from Wikipedia's official statistics~\footnote{The statistics are extracted from {
\url{http://meta.wikimedia.org/wiki/List_of_Wikipedias}}}. We also marked the language analysis applied on the index. Some of the languages in \ourdataset are low-resource ones, especially the ones from the Balto-Slavic family, which is also clear from their Wikipedia sizes. In the table, we see that half of the languages have under one million articles, and Albanian even falls under 100K. Moreover, even more languages are comparable with the number of sentences in the ARC Corpus, which is also built from science books. Finally, some of the languages (Serbian, Croatian, Macedonian, and Albanian) are not processed with any language-specific ElasticSearch analyzers.

\section{Fine-Grained Evaluation}
\label{sec:appendix-fine-grained-eval}

Figure~\ref{fig:finegrained-othersocial} shows fine-grained evaluation for two subject groups: \textit{Social Science} and \textit{Others}. We can see that these subjects are less challenging than Natural Science. One reason is that many of the subjects in these two groups such as Business \& Economics, Geography, and History can be answered using knowledge that is easily accessible in sources such as Wikipedia (e.g., ``\emph{Who was the first prime minister of Poland after 1990?}''), i.e.,~without the need for complex reasoning or calculations, which are often needed in order to answer questions in subjects such as Physics and Chemistry. Nevertheless, while seeing scores as high as 60\% for some subjects and languages, the current multilingual QA models are still far from perfect, which leaves a lot of room for improvement.

\end{document}